\pdfoutput=1

\documentclass[11pt]{article}

\usepackage[preprint]{acl}

\usepackage{times}
\usepackage{latexsym}

\usepackage[T1]{fontenc}

\usepackage[utf8]{inputenc}

\usepackage{microtype}

\usepackage{inconsolata}

\usepackage{graphicx}
\usepackage{booktabs}
\usepackage{adjustbox}
\usepackage{multirow}
\usepackage{subcaption}

\usepackage{enumitem}

\usepackage{amsfonts}
\usepackage{amsmath}
\usepackage{amsthm}
\usepackage{todonotes}
\usepackage{xspace}
\usepackage{longtable}

\usepackage{cleveref}
\crefname{table}{Table}{}
\crefname{figure}{Figure}{Figures}
\crefname{section}{\S}{}
\crefname{appendix}{Appendix}{Appendices}
\theoremstyle{definition}
\newtheorem{definition}{Definition}[section]

\usepackage[]{xcolor,colortbl}

\newcommand{\reasoning}{code\xspace}
\newcommand{\knowledge}{knowledge QA\xspace}
\newcommand{\Reasoning}{Code\xspace}
\newcommand{\Knowledge}{Knowledge QA\xspace}
\newcommand{\code}{code\xspace}

\title{Compute Optimal Scaling of Skills: Knowledge vs Reasoning}

\author{
  \textbf{Nicholas Roberts\textsuperscript{$\mu\dagger$}} \,
  \textbf{Niladri Chatterji\textsuperscript{$\sigma$}} \,
  \textbf{Sharan Narang\textsuperscript{$\sigma$}} \,
  \textbf{Mike Lewis\textsuperscript{$\sigma$}} \,
  \textbf{Dieuwke Hupkes\textsuperscript{$\sigma$}}
\\
\\
  \textsuperscript{$\mu$}University of Wisconsin \,
  \textsuperscript{$\sigma$}GenAI at Meta
\\
\small{
    \textsuperscript{$\dagger$}Work done during an internship at Meta. 
  }
  \\
  \small{
    \textbf{Correspondence:} \, \href{mailto:nick11roberts@cs.wisc.edu}{nick11roberts@cs.wisc.edu} \, \href{mailto:dieuwkehupkes@meta.com}{dieuwkehupkes@meta.com}
    }
}

\begin{document}
\maketitle

\begin{abstract}
Scaling laws are a critical component of the LLM development pipeline, most famously as a way to forecast training decisions such as `compute-optimally' trading-off parameter count and dataset size, alongside a more recent growing list of other crucial decisions. 
In this work, we ask whether compute-optimal scaling behaviour can be skill-dependent.
In particular, we examine knowledge and reasoning-based skills such as knowledge-based QA and code generation, and we answer this question in the affirmative: scaling laws are skill-dependent. 
Next, to understand whether skill-dependent scaling is an artefact of the pretraining datamix, we conduct an extensive ablation of different datamixes and find that, also when correcting for datamix differences, knowledge and code exhibit fundamental differences in scaling behaviour.
We conclude with an analysis of how our findings relate to standard compute-optimal scaling using a validation set, and find that a misspecified validation set can impact compute-optimal parameter count by nearly 50\%, depending on its skill composition. 
\end{abstract}

\begin{figure*}[h]
    \centering
    \begin{subfigure}[b]{0.11\textwidth}
            \includegraphics[height=3.9cm]{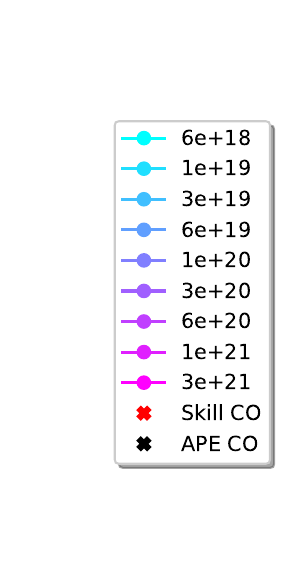}
            \vspace{1mm}
    \end{subfigure}
    \begin{subfigure}[t]{0.29\textwidth}
        \includegraphics[height=4.8cm]{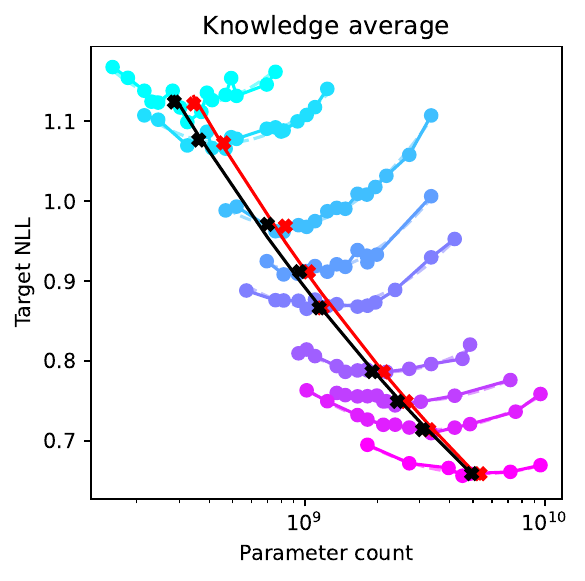}
        \caption{Knowledge QA.}
        \label{fig:canonical_iso_knowledge}
    \end{subfigure}
    \begin{subfigure}[t]{0.29\textwidth}
        \includegraphics[height=4.8cm]{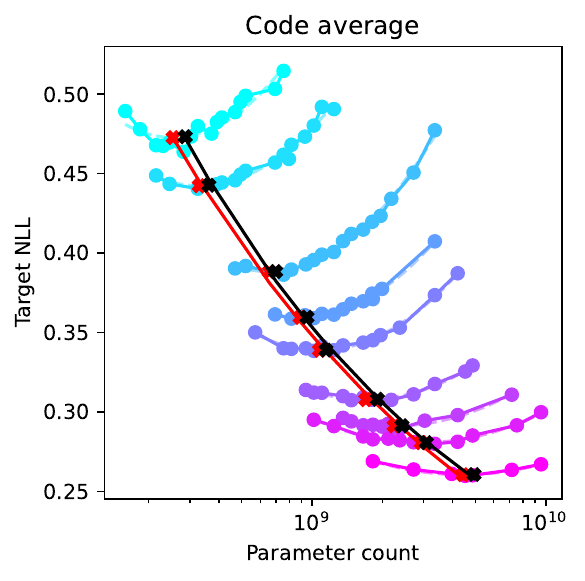}
        \caption{Code.}
        \label{fig:canonical_iso_reasoning}
    \end{subfigure}
    \begin{subfigure}[t]{0.29\textwidth}
        \includegraphics[height=4.8cm]{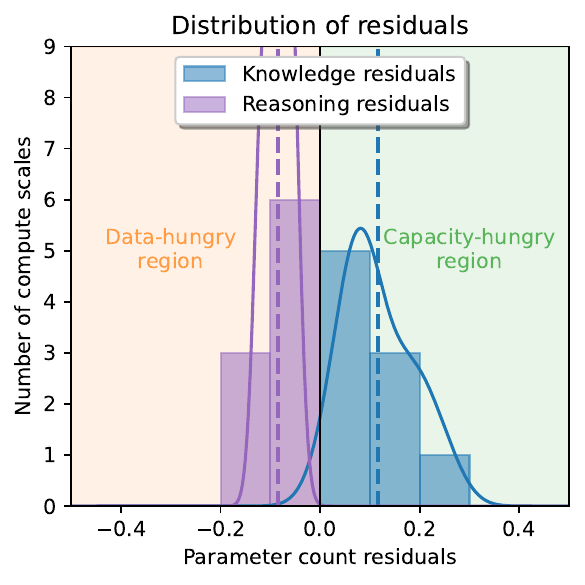}
        \caption{Data- vs capacity-hunger.}
        \label{fig:datahunger_dist}
    \end{subfigure}
    \caption{
        \textbf{Isoflop curves and COs for \reasoning and \knowledge, along with the APE COs.}
        In (a), we see that on average across held-out datasets, \knowledge tends to be capacity-hungry compared to the APE CO. On the other hand, in (b), we see that \code tends to be data-hungry relative to the APE CO. We show the distribution of these relationships in (c), where we plot distributions of the log-scale differences in parameter count between skill-dependent COs and APE COs and find their means lie on opposite sides of the APE CO. 
        The black curves in (a) and (b) represent the predicted APE COs from \citet{llama3}, mapped to their respective IsoFLOP groups. 
    }\label{fig:canonical_iso_all}
\end{figure*}

\section{Introduction}

Used both to forecast performance for early pretraining decisions as well as decide on the optimal trade-off between parameters and pretraining dataset size given a particular compute budget, \emph{scaling laws} \citep[][i.a.]{kaplan2020scaling} have played an important role in the development of large language models (LLMs).
Famously, with a series of experiments with models with different data/parameter trade-offs, \citet{chinchilla} 
showed that previous LLMs were all erring on the side of too many parameters, causing a shift in the amount of training tokens to train LLMs. 
More recently, \citet{llama3} used scaling laws not only to determine the optimal parameter count given their available compute budget, but also to forecast the impact of data selection decisions on evaluation scores.

In these works, the \emph{compute optima} (COs), describing the optimal parameter count and number of training tokens, are selected based on \emph{aggregate performance estimators} (APEs), in the form of negative log-likelihood (NLL) on a validation set not part of the pretraining corpus.
Little is known, however, about whether the COs of individual skills such as mathematical reasoning, question answering (QA), or coding, align with these APE COs.
While some studies use scaling laws to predict how downstream task performance improves with scale \citep[e.g.][]{ye2025data,held2025optimizingpretrainingdatamixtures}, none of these studies cover whether COs themselves may be skill dependent.
Is it possible that some skills are more \emph{data-hungry}, whereas others benefit more from \emph{extra parameters}?
If so, how should that impact model training and training data selection?

In this paper, with an extensive set of experiments across 9 different compute scales and 2 skills as measured with 19 datasets across two different splits, we study exactly that.
Specifically, we focus on the three research questions:

\textbf{R1. Are COs skill dependent?}
First, we consider how the IsoFLOP curves and corresponding COs for \emph{code}\footnote{We use code as a proxy for reasoning skills to avoid ambiguity and the challenge of concretely defining reasoning.}- and \emph{knowledge}-based skills compare to the APE COs, given a canonical datamix. 
Across the board, we find pronounced differences between the COs for these different skills: where \knowledge tasks are capacity-hungry, \code tasks instead prefer data.

\textbf{R2. Is this an artefact of the pretraining datamix, or are code and knowledge skills fundamentally different?}
Using only the smallest compute scale, we next investigate whether these difference are a consequence of the proportion of skill-relevant data in the pretraining datamix of our experiments or that there is in fact a difference in how data- or capacity-hungry the different skills are.
We find that both are true: changing the proportion of skill-relevant data shifts the CO for that skill, but COs for skills differ even with comparable proportions of skill specific data.

\textbf{R3. How does the existence of skill-dependent COs impact LLM training?}
Lastly, we focus on the more practical question of how these findings should impact model pretraining.
First, again using only the smallest compute scale, we investigate if it is possible to \emph{align} the COs for the two skills under consideration. 
We find that it is, but that whether this should be considered advantageous for overall model training depends highly on the choice of validation set.
We then investigate how much the estimated optimal parameter count depends on the choice of validation set.
We find that, on our smallest scale experiments ($6\times10^{18}$), the optimal parameter count for the same skill varies by almost 50\% across datamixes, and by more than 30\% across validation sets.
At larger scale, the difference is smaller, but it still exceeds 10\% for the three largest compute scales.
These experiments show that choosing a validation set that adequately represents what the final model should capture is critical to finding the right parameter counts.

\vspace{-2mm}
\paragraph{Outline} With our work, we contribute to an empirical understanding of how different skills behave across model scales, and explore the concrete implications for CO training of LLMs.
In the remainder of this paper, we first provide the necessary background and definitions to understand our work (\cref{sec:background}), and we describe the data we use for our experiments (\cref{sec:data}).
Next, in \cref{sec:results}, we describe our experiments and their results.
Lastly, we discuss related works about scaling and data selection (\cref{sec:related}), and we conclude in \cref{sec:conclusion}.

\section{Background}
\label{sec:background}

When computing scaling laws, prior work typically focused on modelling the loss on a validation set, which is usually a general text corpus drawn from the same distribution as the pretraining dataset.
In this setup, the loss is effectively a weighted combination of all what the model should learn during training. %
\textit{In contrast, we consider skill-dependent scaling laws.} 
In this section, we discuss the relevant background and notions of scaling laws, COs, and skills underpinning our analysis. 
In \cref{sec:related}, we describe related efforts in more detail, and in \cref{app:definitions}, we provide formal definitions of the concepts that this section introduces. 

\subsection{Scaling laws}
\label{subsec:isoflop_curves}
Neural scaling laws aim to model and predict loss values as a function of the compute scale in FLOPs. 
FLOPs are often estimated as $B \approx 6 p t$, which involves contributions from the size of the training set in tokens, $t$, and the parameter count of the model, $p$. 
Typically, scaling laws that model these two quantities in relation to the loss exhibit a tradeoff between parameter count and compute scale.
These are depicted using \emph{IsoFLOP curves,} where the x-axis is the parameter count, and the y-axis is the loss on the training or validation set. 
To obtain those curves, \citet{chinchilla} pioneered using a 2D power law model which has separate power law components for parameter and token counts. 
In our case, we found 2D power law models to produce a poor fit to downstream evaluations, which we attribute to noise and small dataset sizes, compared to validation sets. 
Instead, we opted to follow the approach used by \citet{llama3}, which involves fitting separate degree-2 polynomials to each compute scale. 
We extend this by fitting a power law to the optima of the compute scales. 

\paragraph{Compute budget and COs}
Central to the idea of IsoFLOP curves, which model the loss at many compute scales, is the idea of \emph{IsoFLOP groups} which model tradeoffs between dataset and model size \emph{at fixed compute scales}. 
In other words, an IsoFLOP group is a set of models where the amount of training data, $t$, and model size, $p$, is varied subject to a fixed approximate compute budget $B \approx 6 p t$. 
This tradeoff between dataset size and parameter count can be optimised at each compute scale, which means that at a given compute budget, there is a \textit{optimal} parameter count and dataset size, $(p^*, t^*) \text{ s.t. } B$ FLOPs. 
Typically, IsoFLOP curves (and therefore COs) are computed using the loss on the training set, or for flexibility, a validation set. 
Since both the training and validation sets include mixtures of data that might influence scaling behaviours differently, we refer to the losses on these sets as \emph{aggregate performance estimators, or APEs}. 
Under this terminology, standard practice involves selecting COs based on APEs. 
Next, we define our notion of skills and how our analysis involves COs based on skills instead of APEs.

\subsection{Skills}
Following \citet{skillit}, we formalise the notion of a `skill' by starting from the metric and dataset by which it can be quantified. 
More precisely, we say that a dataset $\mathcal{D}$ quantifies skill $s$ if the performance of a model on that dataset according to metric $\mathcal{L}$ correlates with a model's ability to perform the intended skill $s$. 
Under this definition, multiple datasets can be associated with the same skill -- something that we exploit in our experiments to validate our conclusions for one skill across multiple datasets -- and in some cases one dataset can quantify multiple skills.
It is difficult to exactly label what skills different datasets quantify, and the extent to which they correctly do so can be open for debate, so we largely follow the labelling of the creators of our evaluation datasets, but we do so under scrutiny. 
For simplicity, our work focuses on two specific skills which we suspect to have different scaling properties: \emph{knowledge}, in the form of knowledge-based QA and \emph{code}.
In \cref{sec:data}, we discuss which datasets we use to quantify these skills.

\paragraph{Skill-dependent COs}
Given the given definition, the CO for a skill $s$ at a particular compute scale $B$ is given by the parameter count $p^*_s$ and training token budget $t^*_s$ in the IsoFLOP group that optimises the loss on the dataset $\mathcal{D}_s$ quantifying the skill, rather than on APEs. 
As mentioned before, a primary aim of our work is to determine whether COs may be skill dependent.
To quantify this, we consider how the COs for skills compare to the optima computed using APEs. 
If a skill fares better with comparatively more \emph{data}, we call this skill \emph{data-hungry}, while if a skill prefers more parameters than the APE optimum we call it \emph{capacity-hungry}.

\begin{table*}[t]

\centering
\begin{adjustbox}{max width=\textwidth}
\begin{tabular}{@{}r|ccccc|ccccc@{}}
\midrule
    \textbf{Split} & \multicolumn{5}{c}{\textbf{Knowledge skills}} & \multicolumn{5}{c}{\textbf{Code skills}} \\
\midrule
\rowcolor[gray]{.9} 
&  \multicolumn{2}{c}{Trivia QA}  & NQ & SQuAD & MMLU & & & CAT-LM & MultiPL-E & \\
\rowcolor[gray]{.9} 
\multirow{-2}{*}{\textbf{Hypothesis}} & \multicolumn{2}{c}{(dev)}  & (dev) & (train) & (train) &
\multirow{-2}{*}{RepoBench} & \multirow{-2}{*}{BigCodeBench} & (test gen.) & (HumanEval) & \multirow{-2}{*}{MBXP}   \\
\multirow{2}{*}{\textbf{Held-out}} & Trivia QA & Trivia QA  & NQ & SQuAD & MMLU &
SWE-bench & BigCodeBench & & \multirow{2}{*}{HumanEval} & \multirow{2}{*}{MBPP} \\
\multirow{-2}{*}{\textbf{Held-out}} & (test) & (wiki test) &  (test) & (dev) & (val) & (oracle) & (test gen.) &  \multirow{-2}{*}{CrossCodeEval} & &   \\
\bottomrule
\end{tabular}
\end{adjustbox}

\caption{
    \textbf{Hypothesis and held-out splits for code and knowledge-based skills.} In all our experiments, we first develop hypothesis on a \emph{hypothesis split} and then confirm them on a \emph{held-out} split containing different datasets for the same skills. References to each of these datasets are provided in the text. 
}
\label{table:trainvalsplit}
\end{table*}

\section{Data}
\label{sec:data}

In our experiments, we focus on two overarching skills: \knowledge and \reasoning.
In this section, we describe which datasets we use to measure models' abilities on those skills and describe how we infer their corresponding COs.

\subsection{Skill data}
For each of the two skills we evaluate, we identify a number of evaluation datasets designed to evaluate these respective skills.
We split those datasets into a `hypothesis' and `held-out' split.
In our analyses, we first form hypotheses on the hypothesis split of \knowledge and \reasoning skills without accessing the held-out split, and then we validate our hypotheses on the held-out split. 
Our hypothesis and held-out splits include \knowledge-based splits from Trivia QA~\citep{joshi-etal-2017-triviaqa}, NQ~\citep{kwiatkowski-etal-2019-natural}, SQuAD~\citep{rajpurkar-etal-2016-squad}, MMLU~\citep{hendryckstest2021}, and \code-based splits from RepoBench~\citep{liu2024repobench}, SWE-Bench~\citep{jimenez2024swebench}, BigCodeBench~\citep{zhuo2025bigcodebench}, CAT-LM~\citep{catlm}, CrossCodeEval~\citep{ding2023crosscodeeval}, MultiPL-E~\citep{multiple}, HumanEval~\citep{chen2021codex}, MBXP~\citep{mbxp_athiwaratkun2022}, and MBPP~\citep{austin2021program}. 
Additional details of the the datasets and splits that we use in our experiments can be found in \cref{table:trainvalsplit}. 

Some of our datasets consist of compound datasets spanning multiple topics, such as MMLU.
We categorise the subtasks of these datasets separately and then assign half of each to the hypothesis and held-out splits, respectively.
We label a subtask as being knowledge-based if the model would have required prior knowledge of the subject area in order to answer the question accurately and remove all non-knowledge-based subtasks. 

\subsection{Pretraining data}
Our canonical pretraining datamix comprises roughly 58.4\% documents that are high in factual knowledge, 19.9\% documents containing code, and the remaining 21.7\% did not fall under either of these categories. 
To construct our training datasets for a given token budget $t$, we randomly sample documents from a larger pool of data from each of these categories, according to the proportions for each category. 
We vary these proportions in \cref{subsec:results2} and \cref{subsec:results_practical} by increasing the proportions of code or by increasing the proportion of knowledge. 
We include visualisations of the proportions for our canonical datamix, along with the proportions used in \cref{subsec:results2} and \cref{subsec:results_practical}, in \cref{app:data_ablations_additional}.

\section{Experiments}
\label{sec:results}

Now, we present the main results for our experiments, starting from our finding that COs are skill specific for our canonical datamix (\cref{subsec:results_skill_cos}), followed by a deeper exploration into the impact of the datamix on these conclusions (\cref{subsec:results2}), and finishing with a small investigation of how these results may impact model pretraining decisions (\cref{subsec:results_practical}).
 
\subsection{Can COs differ between skills?}
\label{subsec:results_skill_cos}

First, we consider the difference in COs for different skills for our canonical datamix.

\paragraph{Methodology}
Roughly following the approach used by \citet{llama3}, we train models for compute budget between $6\times10^{18}$ and $3\times10^{21}$ FLOPs.
At each compute scale, we pretrain models ranging in size between 40M and 8B parameters.\footnote{
Details about model training can be found in \cref{app:model_pretraining}.
}
Using these models, we compute IsoFLOP curves and corresponding COs for the APE (validation loss) and the two skills under consideration -- as measured by NLL of the target answer -- using the methodology described in \cref{subsec:isoflop_curves}. %
Here, in order to compare APE COs with the COs for skills, we first obtain the APE CO parameter counts, $p_c$ from the power law fit given by \citet{llama3}, for each compute scale. 
Next, for a given skill $s$, for each dataset $\mathcal{D}$ belonging to that skill, we fit degree-2 polynomials for each compute scale and identify optimal parameter counts for $p_s$. 
We then project the APE COs onto the polynomial fits for $s$, and fit power law curves to the APE COs and the skill-dependent COs for $s$. 
In all our experiments, we first analyse the results on our hypothesis split and confirm our findings on the held-out split.

\begin{figure*}[t]
    \begin{subfigure}{0.45\textwidth}
    	\centering
        \includegraphics[height=3.0cm]{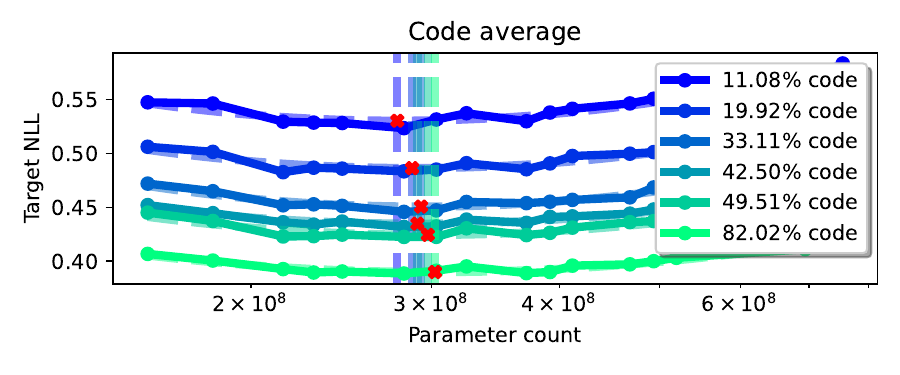}
        \caption{\Reasoning ablation experiments, \reasoning mix.}
        \label{fig:datamix_shift_code}
    \end{subfigure}
    \begin{subfigure}{0.54\textwidth}
        	\centering
        \includegraphics[height=3.0cm]{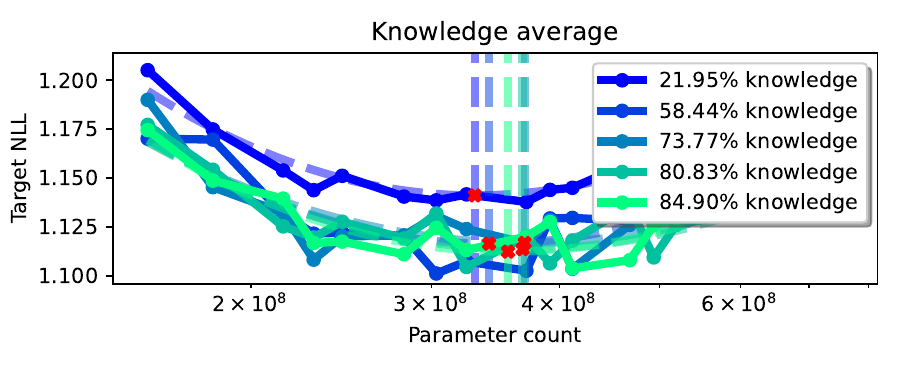}
        \caption{\Knowledge ablation experiments, \knowledge mix.}
        \label{fig:datamix_shift_knowledge}
    \end{subfigure}
    \caption{
    \textbf{$\boldsymbol{6\times10^{18}}$ IsoFLOP curves for various \code and \knowledge datamixes.}
    In (a), we scale the proportion of code pretraining data from $\sim11\%$ to $\sim82\%$ and see the losses improve and the COs shift toward capacity-hunger. 
    For (b), we scale the proportion of knowledge from $\sim22\%$ to $\sim85\%$ and while knowledge tasks appear to be noisier than \code, losses improve and COs shift. 
    }
    \label{fig:datamix_average}
\end{figure*}

\paragraph{Results}
In \cref{fig:canonical_iso_knowledge,fig:canonical_iso_reasoning}, respectively, we show average results for the held-out \reasoning and \knowledge datasets (in red).
For comparison, we overlay the APE CO curve reported by \citet{llama3} (in black).
The figure shows a clear difference between \knowledge and \reasoning: while \knowledge prefers capacity compared to the APE curve -- especially at lower compute scales -- \reasoning tends to be more data-hungry.
This pattern is present across all datasets belonging to the hypothesis split on which we first observed this difference, as well as for the held-out split.
We show individual plots for all datasets in \cref{fig:canonical_iso_knowledge_all_dev,fig:canonical_iso_knowledge_all_heldout,fig:canonical_iso_reasoning_all_dev,fig:canonical_iso_reasoning_all_heldout}.

To make the pattern more explicit across different scales, we consider the distribution of parameter count `residuals,' or log-scale differences between the APE optima and the skill-optima for each skill.
If at compute scale $B$, the residual is larger than 0, we say that the skill is capacity-hungry at $B$; if it is smaller than 0, we call it data-hungry at $B$. 
The resulting distributions, shown in \cref{fig:datahunger_dist}, confirm the pattern we observed before: the parameter count residuals for \reasoning are shifted substantially to the left with respect to the parameter count residuals for \knowledge, indicating that the latter skill is more capacity-hungry.
\textbf{The same pattern holds for all datasets in the hypothesis and held-out splits. This strongly supports the claim that there is a difference in COs between \knowledge and \reasoning, and shows that this observation is a generalisable phenomenon. }

\subsection{Is this an artifact of the data distribution? } 
\label{subsec:results2}
For our canonical datamix, we have seen a clear difference between the COs of \knowledge and \reasoning datasets in their respective capacity- or data-hunger.
What is not clear from our first experiments is whether these dissimilarities are the result of a \textit{fundamental difference between the skills,} or if there is a disparity between the amount of data present for these skills in the canonical datamix. 

\paragraph{Methodology}
To begin to address this question, we first investigate if we can \emph{shift} the skill COs by changing the proportion of relevant data for that skill, which are identified heuristically by tagging the relevant data sources in the pretraining data. 
Starting from the proportion of our canonical dataset (see \cref{fig:canonical_datamix_proportions}), we create four additional datasets for knowledge and five additional datasets for \reasoning in which we up- or downsample the proportion of skill relevant data. 
Specifically, for knowledge, we scale the original proportion of$~58\%$ down to$~22\%$ and up to$~85\%$. 
Similarly for \reasoning, we scale the code proportion from their original$~20\%$ down to$~11\%$ and up to$~84\%$.
With these new datamixes, we train 16 models each with a budget of $6\times10^{18}$ FLOPs, which is the smallest compute scale represented in our IsoFLOP curves, and recompute the skill CO for that compute scale. 

\paragraph{Results}
In \cref{fig:datamix_average}, we show the results for the code and knowledge held-out datasets.
Unsurprisingly, increasing the amount of skill-dependent data (lighter colors in the plot), improves performance for that specific skill.\footnote{Increasing \code data improved losses on \code datasets \emph{beyond the next compute scale on the canonical datamix.}}
This pattern is much clearer for \reasoning than for knowledge.
We hypothesise that this is because code datapoints are easily identified and contain only code, whereas knowledge datasets are usually more dispersed as well as more difficult to label automatically. 
More relevant to our research questions \textit{is the fact that not only the losses, but also the COs shift.}
Again specifically for \reasoning, this pattern is clear in both the hypothesis and held-out sets -- the more code data is included, the more capacity-hungry the CO becomes. 
For reference, in \cref{app:data_ablations_additional}, we provide individual plots for all \reasoning and \knowledge datasets of all splits, across all ablations.
\textbf{This implies a fundamental relationship between the skill proportions in the datamix and COs -- for a fixed compute budget, if the data requirement for a skill is satisfied by increasing its proportion in the datamix, we can afford to train a larger model on less data.}

\begin{figure}[t!]
    \centering
    \includegraphics[width=0.99\columnwidth, trim=0mm 0mm 0mm 0mm, clip]{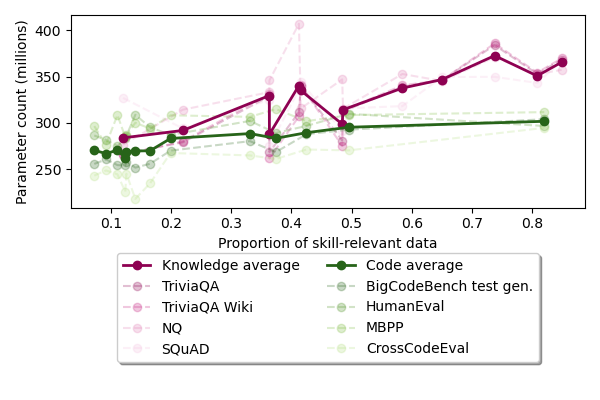}
    \caption{
        \textbf{Optimal parameter count for $\boldsymbol{6\times10^{18}}$ models as a function of proportion of skill-relevant data.}
    When the proportion of skill-relevant data is increased, both knowledge and code skills become more capacity-hungry. 
    On average, knowledge-based tasks require more parameters than code given a particular proportion of skill relevant data, and the number of optimal parameters increases more quickly, implying a fundamental difference between the two skills.
    }
    \label{fig:label_vs_proportion}
\end{figure}

However, we would like to assess whether or not the capacity- and data-hunger of knowledge and code are an artefact of the canonical datamix or if it is a more fundamental phenomenon. 
To do so, we use the same data to investigate the data-hunger of \knowledge and \reasoning \emph{independently} of the proportion of skill relevant data.
To do so, we plot the CO parameter counts for each skill as a function of the proportion of skill relevant dta. 
In \cref{fig:label_vs_proportion}, we show this for both \knowledge and \code in the same plot.\footnote{The axis is aligned on proportions of skill-relevant data, not on the datamixes themselves. For \knowledge datasets (purple lines), an x-value of 0.4 corresponds to a datamix with 40\% knowledge, whereas for \code datasets (green lines), it corresponds to results for the datamix with 40\% \emph{code}.}
From this figure it is clear that even correcting for the proportion of skill relevant data, knowledge is substantially more capacity-hungry than code, and increases its capacity-hunger more quickly as the proportion of skill relevant data increases. 
We hypothesise that this phenomenon, \emph{which demonstrates a fundamental difference between \knowledge and \reasoning}, is related to the fact that knowledge is harder to compress than code, requiring more capacity to memorise facts. 
\textbf{The observation that COs for \knowledge shift at a faster rate than \code suggests a fundamental difference in their scaling, rather than an artefact of the datamix.}

\begin{figure}[t!]
    \centering
    \includegraphics[width=0.99\columnwidth, trim=0mm 0mm 0mm 0mm, clip]{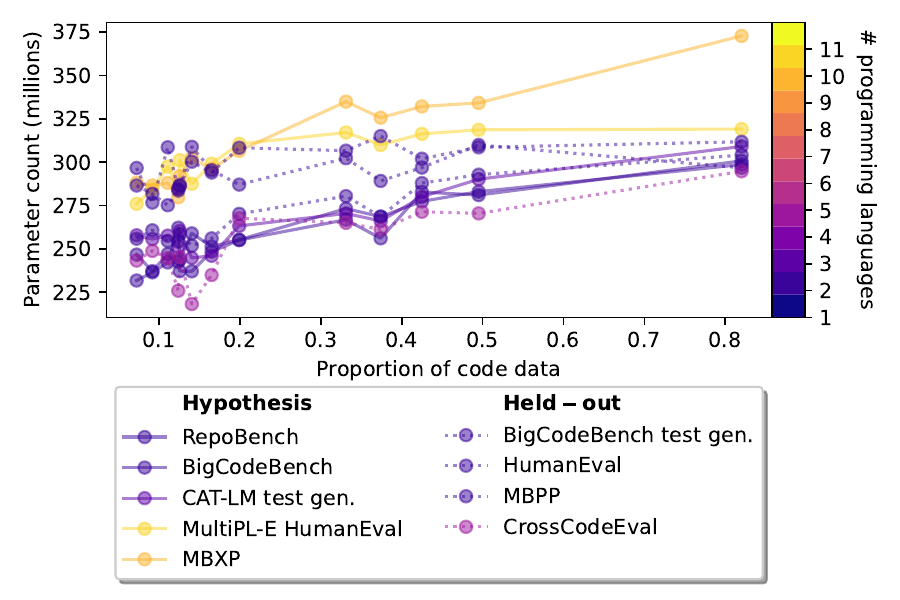}
    \caption{\textbf{Optimal parameters for different coding benchmarks.} Coding skills that contain more and thus less frequent programming languages, shown in yellow, appear to become more capacity-hungry, across all coding skills in our hypothesis and held-out splits.}
    \label{fig:programming_languages}
\end{figure}

In line with this hypothesis, we observe a larger difference between knowledge and code when only Python code datasets are considered, as can be seen in \cref{fig:programming_languages}. 
We hypothesise that models at this scale may not be able to successfully compress `low-resource' programming languages, though further research is required to confirm this.\footnote{
    We also see in \cref{fig:datamix_average,fig:canonical_iso_all} and \cref{app:data_ablations_additional,app:per_benchmark_iso} that datasets quantifying \code have a lower loss value than those quantifying \knowledge -- across the board -- despite large proportions of knowledge, which could suggest that code has \emph{lower unmodelable entropy} compared to knowledge. 
} 

\subsection{What does this imply for LLM training?}\label{subsec:results_practical}
Having empirically shown and investigated the existence of skill-dependent COs, we now explore the practical implications of these findings for pretraining LLMs.
While the research questions that could be explored in this realm are numerous, we focus on two concrete questions that can be feasibly tested within our compute budget:\begin{enumerate}
    \item Can we align COs of skills via data selection to improve the validation loss? 
    \item What is the parameter count impact of a validation set measuring the wrong skills?
\end{enumerate}

\begin{figure*}[t]
    \centering
    \begin{subfigure}[t]{0.25\textwidth}
            \includegraphics[height=2.6cm]{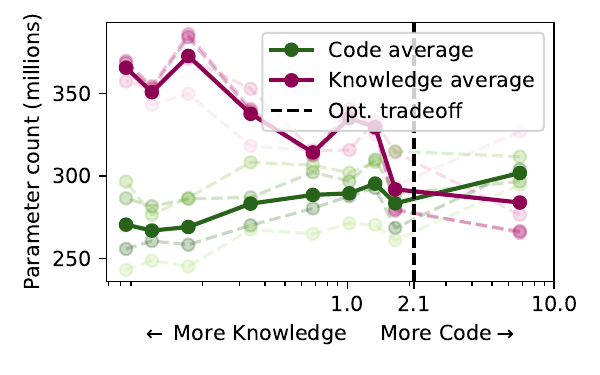}
    \caption{COs alignment.}\label{fig:crossover_kc}
    \end{subfigure}
    \begin{subfigure}[t]{0.74\textwidth}
            \includegraphics[height=2.75cm]{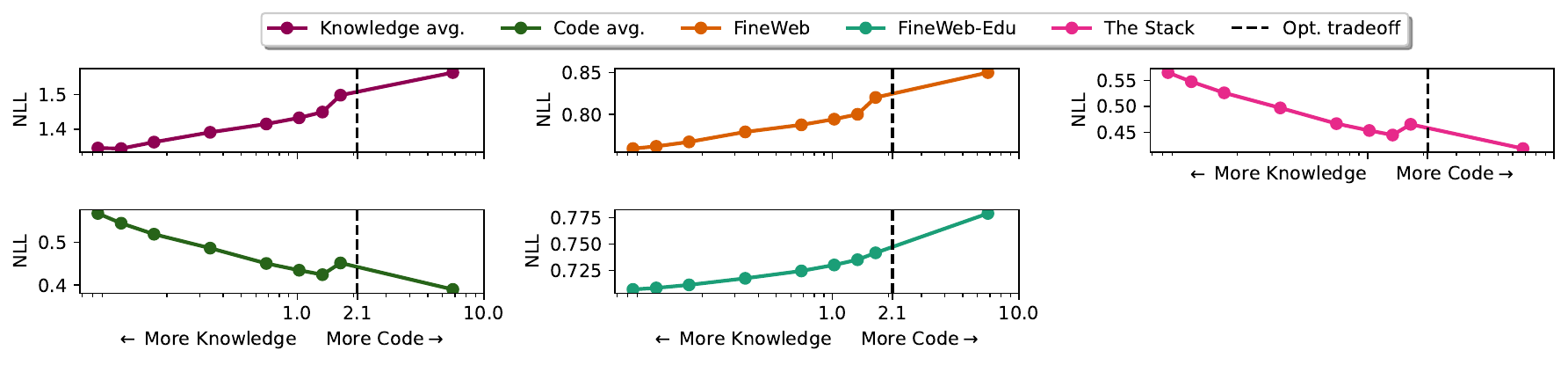}
              \caption{The impact of COs alignment on \knowledge, \code, and validation sets.}\label{fig:crossover_nll}
    \end{subfigure}
    \caption{\textbf{Optimal parameter count and loss as a function of code/knowledge ratio.}
        (a) Optimal parameter count for $6\times10^{18}$ models, as a function of the ratio between code and knowledge in the pretraining data. Dashed lines indicate the `crossover point,' indicating the ratio for which the optimal parameter counts align. 
        (b) Evaluation and validation loss for the same models, with the same optimal cross-over point. 
        Every validation set that we consider exhibits losses that closely follow \emph{either} the average \knowledge or average \code curves. 
        }\label{fig:crossover_kc_nll}
\end{figure*}

\subsubsection{Aligning skill-COs}
Given that COs of knowledge-related skills are more capacity-hungry while code related skills instead benefit more from data, we now ask the question if we can improve the overall aggregate performance of a model by shifting the datamix such that the COs of the two skills are aligned.
In \cref{fig:crossover_kc}, we plot the optimal parameter count for the $6\times10^{18}$ models we trained for the results reported in the previous subsection.
\textbf{We can see that there is indeed a point where the optimal parameter counts for code and knowledge roughly match, which is when code data is approximately 2.1 times as prevalent as knowledge data.}\footnote{We obtain this value of 2.1 by computing the crossover point on the hypothesis set and overlaying it onto the held-out splits -- we include a comparison between the crossover points found on the hypothesis and held-out splits in \cref{app:crossover_comp}.} 
This is in line with our earlier results: as we add more code data, the CO for code will very gradually shift towards being more parameter-hungry, whereas decreasing the amount of knowledge data shifts the CO for knowledge to require fewer parameters.

Next, we look at how this trade-off impacts the validation loss, indicative of the aggregate model performance. 
Since our original APE COs were projected from the scaling law parameters given by~\citet{llama3}, we would ideally conduct this analysis using the Llama 3 validation set. 
However, details of this validation set are unreleased, so, instead, we sample three different validation datasets from different publicly available pretraining datasets: FineWeb, FineWeb-Edu~\citep{penedo2024the} comprising educational webpages filtered from FineWeb, and The Stack~\cite{Kocetkov2022TheStack}, a \code-based pretraining dataset. 
We randomly draw 20K samples from each of these datasets to produce a much smaller validation set. 
For FineWeb\footnote{\scriptsize\url{https://huggingface.co/datasets/HuggingFaceFW/fineweb}} and FineWeb-Edu,\footnote{\scriptsize\url{https://huggingface.co/datasets/HuggingFaceFW/fineweb-edu}} we draw the samples from their 10B token sample splits.
For The Stack, we draw the samples from the subsampled The Stack Smol, containing a small subset of approximately $0.1\%$ of The Stack.\footnote{\scriptsize\url{https://huggingface.co/datasets/bigcode/the-stack-smol}}  

In \cref{fig:crossover_nll}, we show the NLL for all validation sets, using the same axis as in \cref{fig:crossover_kc}. 
The vertical dashed line indicates the optimal point from \cref{fig:crossover_kc}. 
A first striking observation is that all the chosen validation sets track either \knowledge or \code quite closely: while the loss on FineWeb and FineWeb-Edu monotonically increases with the amount of code data, the opposite is true for The Stack.
We do not see evidence of a mix of \knowledge and \code in the validation sets, which might have presented itself as a U-shaped curve, rather than the monotonically increasing or decreasing curves that we observe. 
To compute appropriate scaling laws, none of these validation sets would thus have resulted in COs aligned with \emph{both} skills.
\textbf{This result implies that it is possible to align COs between skills, and that this should be done in practice to balance desired skills.}

\begin{figure*}[h]
    \centering
    \begin{subfigure}[t]{0.24\textwidth}
        \includegraphics[height=3.0cm]{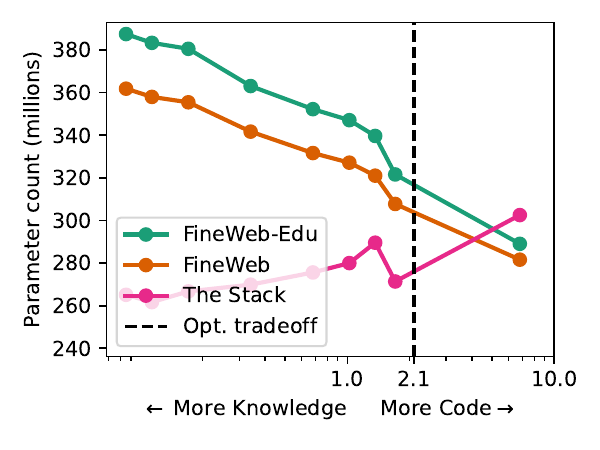}
        \vspace{-6mm}
        \caption{}\label{fig:crossover_val}
    \end{subfigure}
    \begin{subfigure}[t]{0.24\textwidth}
        \includegraphics[height=3.0cm]{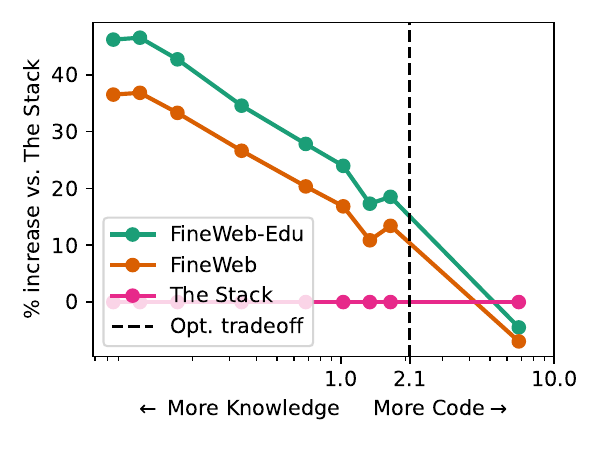}
        \vspace{-6mm}
        \caption{}\label{fig:crossover_val_rel}
    \end{subfigure}
    \begin{subfigure}[t]{0.24\textwidth}
        \includegraphics[height=3.0cm]{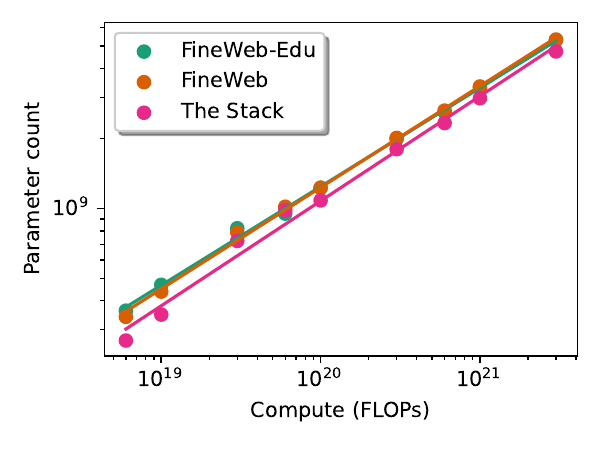}
        \vspace{-6mm}
        \caption{}\label{fig:crossover_val_scaling}
    \end{subfigure}
    \begin{subfigure}[t]{0.24\textwidth}
        \includegraphics[height=3.0cm]{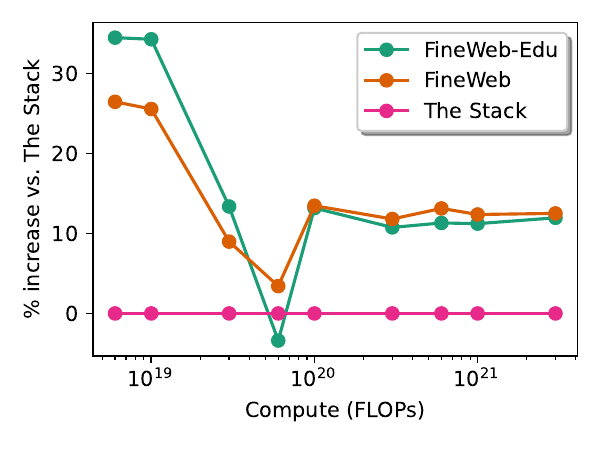}
        \vspace{-6mm}
        \caption{}\label{fig:crossover_val_scaling_rel}
    \end{subfigure}
    \caption{
        \textbf{Fluctuation in optimal parameter count as function of validation set.} In (a), we show the COs of the validation sets that we evaluate, showing that at a high ratio of \knowledge to \code, two of the three validation sets have COs with substantially higher parameter counts. In (b), we see this information in terms of a relative increase in parameter count compared to The Stack, and see that the increase can be up to roughly $50\%$. In (c), we show the CO scaling behaviour across compute scales between validation sets. In (d), we see the relative parameter count increases over The Stack, and see that at high compute scales, the increase is smaller, yet it still exceeds $10\%$. 
    }
\label{fig:optimal_params_val}
\end{figure*}

\subsubsection{The impact of the validation set}
Lastly, we study more explicitly how much variation in the predicted COs is to be expected depending on the choice of validation set.
In \cref{fig:crossover_val}, considering The Stack validation set as `baseline,' we can see that for some datamixes, the optimal parameter count is off by almost 50\%.
In \cref{fig:crossover_val_scaling}, we see that, for the smallest compute scale, for the canonical datamix there is over a 30\% difference in optimal parameter count between the different validation sets. 
While the relative difference is not as stark at higher compute scales, it still exceeds 10\% for the 3 largest compute scales.
\textbf{This experiment shows that choosing a validation set that adequately represents a mix of what the final model should capture -- or perhaps even a validation set that more directly measures those skills, such as in our work -- is critical to finding the right COs.
While the difference is more pronounced at smaller scales, it persists even on the largest compute scale in our experiments.}

On the other hand, \citet{llama3} points out that IsoFLOP curves get flatter at high compute scales, so the tradeoff between model capacity and dataset size might not matter as much. 
In one of their other results, \citet{llama3} uses a fitted sigmoid of the loss to estimate \emph{accuracies}. 
If one were to take their analysis a step further by applying a sigmoid to an entire IsoFLOP curve to produce an IsoFLOP curve of accuracy rather than NLL, we hypothesise that this could make curves appear \emph{less flat} if low compute scales are in the flat region of the sigmoid and high compute scales are in the linear region. 
We leave this intriguing question to future work.

\section{Related work}
\label{sec:related}

\paragraph{Scaling laws and compute optimality}
The first large-scale empirical study of statistical scaling laws of neural networks was done by \citet{hestness2017deeplearningscalingpredictable}, who studied scaling laws for a variety of tasks using the LSTM~\cite{lstm} and CNN~\cite{cnn} architectures. 
Their paper used power law fits to their experimental data, and they found that the only way to materially change the power law exponent was to \textit{change the task}, whereas other choices such as hyperparameters changed only the scaling factor. 
Later works extended these ideas to joint model and dataset size scaling, such as those by \citet{Rosenfeld2019ACP,kaplan2020scaling}, and famously, the Chinchilla paper of~\citet{chinchilla} which introduced the notion of CO scaling measured using the loss on the training set. 
More recently, several efforts aim to predict downstream performance using scaling laws \citep{isik2025scaling,bhagia2024establishingtaskscalinglaws}, but they focus on forecasting downstream metrics, whereas our work asks a fundamental question about how different skills exhibit different scaling behaviour. 
Related to these efforts is a concurrent work that characterises LLM skills as latent variables of the scaling model fit to benchmark scores of existing pretrained models, but differs from our work in its treatment of skills and crucially, that it does not study compute optimality~\cite{polo2025slothscalinglawsllm}. 

\paragraph{Data selection}
In addition to scaling laws, our work is related to the rich literature on data selection and optimising the pretraining datamix. 
Inspired by scaling laws, the recent Data Mixing Laws \cite{ye2025data} aim to model the loss as a function of the proportions of different types of data in the pretraining datamix. 
Other work attempts to improve dataset quality and therefore downstream performance by optimising the datamix \cite{chen2025aioli,xie2023doremi,held2025optimizingpretrainingdatamixtures}, as well as Skill-It \cite{skillit}, which dynamically updates the dataset proportions throughout training using downstream validation losses and an online mirror descent algorithm. 
However, these works do not focus on how COs might shift as a function of the pretraining datamix. 

\paragraph{Skill frameworks}
Skill-It formalises the downstream losses that it targets as `skills,' which we draw inspiration from in our work. 
In Skill-It, a skill is defined as a unit of behaviour associated with a dataset and a metric, such that if a model is trained on samples from the dataset, its metric improves on unseen samples. 
However, there are other related definitions of skills that include more general notions of behaviour such as reasoning abilities \cite{xia2024less,Arora2023ATF} and the evaluation thereof \cite{yu2024skillmix}.

\section{Conclusion}
\label{sec:conclusion}

In this work, motivated by the impact that scaling laws have had on the landscape of training LLMs, we examine the existence of \emph{skill-dependent scaling laws}.
Specifically, we investigate how the optimal trade-off between capacity (in the form of parameter count) and training data varies across skills.
Focusing on two different skills -- code generation and knowledge-based QA -- we find pronounced differences in how \emph{data}- or \emph{capacity-hungry} they are: where \knowledge is capacity-hungry, \code instead prefers data.
These differences hold across tasks and splits, and are true even if the proportion of \emph{skill-dependent data} in the pretraining datamix is factored out.
Furthermore, as the proportion of skill-dependent data grows, the capacity-hunger of knowledge-based skills grows more rapidly than that of code, indicating that models are able to compress the latter better than the former.

In the second part of our experiments, we focus on the impact of these findings to LLM training.
We find that both the use of validation set and the specific datamix used to do scaling experiments have a substantial impact on the estimated compute optimal points.
In our $6\times10^{18}$ FLOP experiments, we find that varying the ratio of code and knowledge data can result in a difference of 30\% or more in the estimated optimal parameter count for that budget for a single validation set, and almost 50\% depending on the validation set.
Similarly, the use of validation set can result in a mismatch of over 30\% on the smallest scale, and over 10\% on the largest three scales (up to $3\times10^{21}$).
As such, our experiments show that choosing a validation set that adequately represents a mix of what the final model should capture is critical to finding the right COs as well as that doing data ablations \emph{after} selecting COs may result in suboptimal parameter counts for the specific datamix.

Our analysis also opens intriguing directions for future study, such as understanding the relationship between compression and skills through the lens of COs, as well as understanding the flatness or curvature of IsoFLOP curves computed on the NLL compared to accuracy-based metrics. 
Crucially, we note that the bulk of our analysis can be easily extended to additional skills using open source scaling suites, as it only requires running evaluations on pretrained models.

\section{Limitations}
\label{sec:limitations}

The primary limitation of our analysis is that our datamix ablation analysis in \cref{subsec:results2} and part of our analysis in \cref{subsec:results_practical} was conducted using only the compute scale of $6 \times 10^{18}$ FLOPs, which is the lowest that was represented in our IsoFLOP curves from \cref{subsec:results_skill_cos}. 
In the ideal case, the same analysis would have been repeated across all of the compute scales that we consider, but this would have been prohibitively expensive, as it would have required training well over 1,000 models at higher compute scales. 
A second limitation is of the sizes of our hypothesis and held-out splits. 
Ideally, these splits would contain a much larger number of benchmarks for \knowledge and \code, but we found that it was challenging to obtain mode \knowledge benchmarks that were large, \knowledge-specific and isolated from reasoning skills, and were sufficiently low in noise at low compute scales. 
Third, we were unable to deeply explore our hypothesis that non-Python languages are underrepresented in the training data and are therefore harder to compress (see \cref{fig:programming_languages} and \cref{subsec:results2}). 
This was because we did not set out to construct the hypothesis and held-out splits in a way that took the number of programming languages into account, so we were unable to rigorously validate this observation using the held-out split. 
Nonetheless, as discussed in \cref{subsec:results2}, we leave this interesting question to future work. 
Finally, while we used \code as a proxy for reasoning skills so as to avoid the challenge of defining reasoning, there may indeed be other types of reasoning skills that are somehow fundamentally different from \code. 
We hope that future work can shed light on what constitutes reasoning, how it relates to \code, and its relationship to other skills.

\section*{Acknowledgements}
\label{sec:acknowledgements}
We thank Lovish Madaan and Frank Zhang for the technical and infrastructure help throughout the project, along with the many helpful discussions. 
We also thank Elia Bruni for his input and suggestions for this project. 
Nicholas Roberts additionally thanks Rishi Hazra and Deepak Nathani for the many Friday-night research chats at the pub, where he learned a lot about their research along with how to explain this work.

\bibliography{refs}

\begin{thebibliography}{34}
\providecommand{\natexlab}[1]{#1}

\bibitem[{Arora and Goyal(2023)}]{Arora2023ATF}
Sanjeev Arora and Anirudh Goyal. 2023.
\newblock \href {https://api.semanticscholar.org/CorpusID:260334352} {A theory
  for emergence of complex skills in language models}.
\newblock \emph{ArXiv}, abs/2307.15936.

\bibitem[{Athiwaratkun et~al.(2022)Athiwaratkun, Gouda, Wang, Li, Tian, Tan,
  Ahmad, Wang, Sun, Shang, Gonugondla, Ding, Kumar, Fulton, Farahani, Jain,
  Giaquinto, Qian, Ramanathan, Nallapati, Ray, Bhatia, Sengupta, Roth, and
  Xiang}]{mbxp_athiwaratkun2022}
Ben Athiwaratkun, Sanjay~Krishna Gouda, Zijian Wang, Xiaopeng Li, Yuchen Tian,
  Ming Tan, Wasi~Uddin Ahmad, Shiqi Wang, Qing Sun, Mingyue Shang, Sujan~Kumar
  Gonugondla, Hantian Ding, Varun Kumar, Nathan Fulton, Arash Farahani,
  Siddhartha Jain, Robert Giaquinto, Haifeng Qian, Murali~Krishna Ramanathan,
  Ramesh Nallapati, Baishakhi Ray, Parminder Bhatia, Sudipta Sengupta, Dan
  Roth, and Bing Xiang. 2022.
\newblock \href {https://doi.org/10.48550/ARXIV.2210.14868} {Multi-lingual
  evaluation of code generation models}.

\bibitem[{Austin et~al.(2021)Austin, Odena, Nye, Bosma, Michalewski, Dohan,
  Jiang, Cai, Terry, Le et~al.}]{austin2021program}
Jacob Austin, Augustus Odena, Maxwell Nye, Maarten Bosma, Henryk Michalewski,
  David Dohan, Ellen Jiang, Carrie Cai, Michael Terry, Quoc Le, et~al. 2021.
\newblock Program synthesis with large language models.
\newblock \emph{arXiv preprint arXiv:2108.07732}.

\bibitem[{Bhagia et~al.(2024)Bhagia, Liu, Wettig, Heineman, Tafjord, Jha,
  Soldaini, Smith, Groeneveld, Koh, Dodge, and
  Hajishirzi}]{bhagia2024establishingtaskscalinglaws}
Akshita Bhagia, Jiacheng Liu, Alexander Wettig, David Heineman, Oyvind Tafjord,
  Ananya~Harsh Jha, Luca Soldaini, Noah~A. Smith, Dirk Groeneveld, Pang~Wei
  Koh, Jesse Dodge, and Hannaneh Hajishirzi. 2024.
\newblock \href {https://arxiv.org/abs/2412.04403} {Establishing task scaling
  laws via compute-efficient model ladders}.
\newblock \emph{Preprint}, arXiv:2412.04403.

\bibitem[{Cassano et~al.(2023)Cassano, Gouwar, Nguyen, Nguyen, Phipps-Costin,
  Pinckney, Yee, Zi, Anderson, Feldman, Guha, Greenberg, and Jangda}]{multiple}
Federico Cassano, John Gouwar, Daniel Nguyen, Sydney Nguyen, Luna
  Phipps-Costin, Donald Pinckney, Ming-Ho Yee, Yangtian Zi, Carolyn~Jane
  Anderson, Molly~Q Feldman, Arjun Guha, Michael Greenberg, and Abhinav Jangda.
  2023.
\newblock \href {https://doi.org/10.1109/TSE.2023.3267446} {Multipl-e: A
  scalable and polyglot approach to benchmarking neural code generation}.
\newblock \emph{IEEE Trans. Softw. Eng.}, 49(7):3675–3691.

\bibitem[{Chen et~al.(2021)Chen, Tworek, Jun, Yuan, de~Oliveira~Pinto, Kaplan,
  Edwards, Burda, Joseph, Brockman, Ray, Puri, Krueger, Petrov, Khlaaf, Sastry,
  Mishkin, Chan, Gray, Ryder, Pavlov, Power, Kaiser, Bavarian, Winter, Tillet,
  Such, Cummings, Plappert, Chantzis, Barnes, Herbert-Voss, Guss, Nichol,
  Paino, Tezak, Tang, Babuschkin, Balaji, Jain, Saunders, Hesse, Carr, Leike,
  Achiam, Misra, Morikawa, Radford, Knight, Brundage, Murati, Mayer, Welinder,
  McGrew, Amodei, McCandlish, Sutskever, and Zaremba}]{chen2021codex}
Mark Chen, Jerry Tworek, Heewoo Jun, Qiming Yuan, Henrique~Ponde
  de~Oliveira~Pinto, Jared Kaplan, Harri Edwards, Yuri Burda, Nicholas Joseph,
  Greg Brockman, Alex Ray, Raul Puri, Gretchen Krueger, Michael Petrov, Heidy
  Khlaaf, Girish Sastry, Pamela Mishkin, Brooke Chan, Scott Gray, Nick Ryder,
  Mikhail Pavlov, Alethea Power, Lukasz Kaiser, Mohammad Bavarian, Clemens
  Winter, Philippe Tillet, Felipe~Petroski Such, Dave Cummings, Matthias
  Plappert, Fotios Chantzis, Elizabeth Barnes, Ariel Herbert-Voss,
  William~Hebgen Guss, Alex Nichol, Alex Paino, Nikolas Tezak, Jie Tang, Igor
  Babuschkin, Suchir Balaji, Shantanu Jain, William Saunders, Christopher
  Hesse, Andrew~N. Carr, Jan Leike, Josh Achiam, Vedant Misra, Evan Morikawa,
  Alec Radford, Matthew Knight, Miles Brundage, Mira Murati, Katie Mayer, Peter
  Welinder, Bob McGrew, Dario Amodei, Sam McCandlish, Ilya Sutskever, and
  Wojciech Zaremba. 2021.
\newblock \href {https://arxiv.org/abs/2107.03374} {Evaluating large language
  models trained on code}.

\bibitem[{Chen et~al.(2023)Chen, Roberts, Bhatia, Wang, Zhang, Sala, and
  R\'{e}}]{skillit}
Mayee Chen, Nicholas Roberts, Kush Bhatia, Jue Wang, Ce~Zhang, Frederic Sala,
  and Christopher R\'{e}. 2023.
\newblock \href
  {https://proceedings.neurips.cc/paper_files/paper/2023/file/70b8505ac79e3e131756f793cd80eb8d-Paper-Conference.pdf}
  {Skill-it! a data-driven skills framework for understanding and training
  language models}.
\newblock In \emph{Advances in Neural Information Processing Systems},
  volume~36, pages 36000--36040. Curran Associates, Inc.

\bibitem[{Chen et~al.(2025)Chen, Hu, Lourie, Cho, and Re}]{chen2025aioli}
Mayee~F Chen, Michael~Y. Hu, Nicholas Lourie, Kyunghyun Cho, and Christopher
  Re. 2025.
\newblock \href {https://openreview.net/forum?id=sZGZJhaNSe} {Aioli: A unified
  optimization framework for language model data mixing}.
\newblock In \emph{The Thirteenth International Conference on Learning
  Representations}.

\bibitem[{Ding et~al.(2023)Ding, Wang, Ahmad, Ding, Tan, Jain, Ramanathan,
  Nallapati, Bhatia, Roth, and Xiang}]{ding2023crosscodeeval}
Yangruibo Ding, Zijian Wang, Wasi~Uddin Ahmad, Hantian Ding, Ming Tan, Nihal
  Jain, Murali~Krishna Ramanathan, Ramesh Nallapati, Parminder Bhatia, Dan
  Roth, and Bing Xiang. 2023.
\newblock \href {https://openreview.net/forum?id=wgDcbBMSfh} {Crosscodeeval: A
  diverse and multilingual benchmark for cross-file code completion}.
\newblock In \emph{Thirty-seventh Conference on Neural Information Processing
  Systems Datasets and Benchmarks Track}.

\bibitem[{Dubey et~al.(2024)Dubey, Jauhri, Pandey, Kadian, Al-Dahle, Letman,
  Mathur, Schelten, Yang, Fan et~al.}]{llama3}
Abhimanyu Dubey, Abhinav Jauhri, Abhinav Pandey, Abhishek Kadian, Ahmad
  Al-Dahle, Aiesha Letman, Akhil Mathur, Alan Schelten, Amy Yang, Angela Fan,
  et~al. 2024.
\newblock The llama 3 herd of models.
\newblock \emph{arXiv preprint arXiv:2407.21783}.

\bibitem[{Held et~al.(2025)Held, Paranjape, Koura, Lewis, Zhang, and
  Mihaylov}]{held2025optimizingpretrainingdatamixtures}
William Held, Bhargavi Paranjape, Punit~Singh Koura, Mike Lewis, Frank Zhang,
  and Todor Mihaylov. 2025.
\newblock \href {https://arxiv.org/abs/2501.11747} {Optimizing pretraining data
  mixtures with llm-estimated utility}.
\newblock \emph{Preprint}, arXiv:2501.11747.

\bibitem[{Hendrycks et~al.(2021)Hendrycks, Burns, Basart, Zou, Mazeika, Song,
  and Steinhardt}]{hendryckstest2021}
Dan Hendrycks, Collin Burns, Steven Basart, Andy Zou, Mantas Mazeika, Dawn
  Song, and Jacob Steinhardt. 2021.
\newblock Measuring massive multitask language understanding.
\newblock \emph{Proceedings of the International Conference on Learning
  Representations (ICLR)}.

\bibitem[{Hestness et~al.(2017)Hestness, Narang, Ardalani, Diamos, Jun,
  Kianinejad, Patwary, Yang, and
  Zhou}]{hestness2017deeplearningscalingpredictable}
Joel Hestness, Sharan Narang, Newsha Ardalani, Gregory Diamos, Heewoo Jun,
  Hassan Kianinejad, Md. Mostofa~Ali Patwary, Yang Yang, and Yanqi Zhou. 2017.
\newblock \href {https://arxiv.org/abs/1712.00409} {Deep learning scaling is
  predictable, empirically}.
\newblock \emph{Preprint}, arXiv:1712.00409.

\bibitem[{Hochreiter and Schmidhuber(1997)}]{lstm}
Sepp Hochreiter and J\"{u}rgen Schmidhuber. 1997.
\newblock \href {https://doi.org/10.1162/neco.1997.9.8.1735} {Long short-term
  memory}.
\newblock \emph{Neural Comput.}, 9(8):1735–1780.

\bibitem[{Hoffmann et~al.(2022)Hoffmann, Borgeaud, Mensch, Buchatskaya, Cai,
  Rutherford, de~Las~Casas, Hendricks, Welbl, Clark, Hennigan, Noland,
  Millican, van~den Driessche, Damoc, Guy, Osindero, Simonyan, Elsen, Vinyals,
  Rae, and Sifre}]{chinchilla}
Jordan Hoffmann, Sebastian Borgeaud, Arthur Mensch, Elena Buchatskaya, Trevor
  Cai, Eliza Rutherford, Diego de~Las~Casas, Lisa~Anne Hendricks, Johannes
  Welbl, Aidan Clark, Tom Hennigan, Eric Noland, Katie Millican, George van~den
  Driessche, Bogdan Damoc, Aurelia Guy, Simon Osindero, Karen Simonyan, Erich
  Elsen, Oriol Vinyals, Jack~W. Rae, and Laurent Sifre. 2022.
\newblock Training compute-optimal large language models.
\newblock In \emph{Proceedings of the 36th International Conference on Neural
  Information Processing Systems}, NIPS '22, Red Hook, NY, USA. Curran
  Associates Inc.

\bibitem[{Isik et~al.(2025)Isik, Ponomareva, Hazimeh, Paparas, Vassilvitskii,
  and Koyejo}]{isik2025scaling}
Berivan Isik, Natalia Ponomareva, Hussein Hazimeh, Dimitris Paparas, Sergei
  Vassilvitskii, and Sanmi Koyejo. 2025.
\newblock \href {https://openreview.net/forum?id=vPOMTkmSiu} {Scaling laws for
  downstream task performance in machine translation}.
\newblock In \emph{The Thirteenth International Conference on Learning
  Representations}.

\bibitem[{Jimenez et~al.(2024)Jimenez, Yang, Wettig, Yao, Pei, Press, and
  Narasimhan}]{jimenez2024swebench}
Carlos~E Jimenez, John Yang, Alexander Wettig, Shunyu Yao, Kexin Pei, Ofir
  Press, and Karthik~R Narasimhan. 2024.
\newblock \href {https://openreview.net/forum?id=VTF8yNQM66} {{SWE}-bench: Can
  language models resolve real-world github issues?}
\newblock In \emph{The Twelfth International Conference on Learning
  Representations}.

\bibitem[{Joshi et~al.(2017)Joshi, Choi, Weld, and
  Zettlemoyer}]{joshi-etal-2017-triviaqa}
Mandar Joshi, Eunsol Choi, Daniel Weld, and Luke Zettlemoyer. 2017.
\newblock \href {https://doi.org/10.18653/v1/P17-1147} {{T}rivia{QA}: A large
  scale distantly supervised challenge dataset for reading comprehension}.
\newblock In \emph{Proceedings of the 55th Annual Meeting of the Association
  for Computational Linguistics (Volume 1: Long Papers)}, pages 1601--1611,
  Vancouver, Canada. Association for Computational Linguistics.

\bibitem[{Kaplan et~al.(2020)Kaplan, McCandlish, Henighan, Brown, Chess, Child,
  Gray, Radford, Wu, and Amodei}]{kaplan2020scaling}
Jared Kaplan, Sam McCandlish, Tom Henighan, Tom~B. Brown, Benjamin Chess, Rewon
  Child, Scott Gray, Alec Radford, Jeffrey Wu, and Dario Amodei. 2020.
\newblock Scaling laws for neural language models.
\newblock \emph{arXiv preprint arXiv:2001.08361}.

\bibitem[{Kocetkov et~al.(2022)Kocetkov, Li, Ben~Allal, Li, Mou,
  Muñoz~Ferrandis, Jernite, Mitchell, Hughes, Wolf, Bahdanau, von Werra, and
  de~Vries}]{Kocetkov2022TheStack}
Denis Kocetkov, Raymond Li, Loubna Ben~Allal, Jia Li, Chenghao Mou, Carlos
  Muñoz~Ferrandis, Yacine Jernite, Margaret Mitchell, Sean Hughes, Thomas
  Wolf, Dzmitry Bahdanau, Leandro von Werra, and Harm de~Vries. 2022.
\newblock The stack: 3 tb of permissively licensed source code.
\newblock \emph{Preprint}.

\bibitem[{Kwiatkowski et~al.(2019)Kwiatkowski, Palomaki, Redfield, Collins,
  Parikh, Alberti, Epstein, Polosukhin, Devlin, Lee, Toutanova, Jones, Kelcey,
  Chang, Dai, Uszkoreit, Le, and Petrov}]{kwiatkowski-etal-2019-natural}
Tom Kwiatkowski, Jennimaria Palomaki, Olivia Redfield, Michael Collins, Ankur
  Parikh, Chris Alberti, Danielle Epstein, Illia Polosukhin, Jacob Devlin,
  Kenton Lee, Kristina Toutanova, Llion Jones, Matthew Kelcey, Ming-Wei Chang,
  Andrew~M. Dai, Jakob Uszkoreit, Quoc Le, and Slav Petrov. 2019.
\newblock \href {https://doi.org/10.1162/tacl_a_00276} {Natural questions: A
  benchmark for question answering research}.
\newblock \emph{Transactions of the Association for Computational Linguistics},
  7:452--466.

\bibitem[{Lecun et~al.(1998)Lecun, Bottou, Bengio, and Haffner}]{cnn}
Y.~Lecun, L.~Bottou, Y.~Bengio, and P.~Haffner. 1998.
\newblock \href {https://doi.org/10.1109/5.726791} {Gradient-based learning
  applied to document recognition}.
\newblock \emph{Proceedings of the IEEE}, 86(11):2278--2324.

\bibitem[{Liu et~al.(2024)Liu, Xu, and McAuley}]{liu2024repobench}
Tianyang Liu, Canwen Xu, and Julian McAuley. 2024.
\newblock \href {https://openreview.net/forum?id=pPjZIOuQuF} {Repobench:
  Benchmarking repository-level code auto-completion systems}.
\newblock In \emph{The Twelfth International Conference on Learning
  Representations}.

\bibitem[{Penedo et~al.(2024)Penedo, Kydl{\'\i}{\v{c}}ek, allal, Lozhkov,
  Mitchell, Raffel, Werra, and Wolf}]{penedo2024the}
Guilherme Penedo, Hynek Kydl{\'\i}{\v{c}}ek, Loubna~Ben allal, Anton Lozhkov,
  Margaret Mitchell, Colin Raffel, Leandro~Von Werra, and Thomas Wolf. 2024.
\newblock \href {https://openreview.net/forum?id=n6SCkn2QaG} {The fineweb
  datasets: Decanting the web for the finest text data at scale}.
\newblock In \emph{The Thirty-eight Conference on Neural Information Processing
  Systems Datasets and Benchmarks Track}.

\bibitem[{Polo et~al.(2025)Polo, Somerstep, Choshen, Sun, and
  Yurochkin}]{polo2025slothscalinglawsllm}
Felipe~Maia Polo, Seamus Somerstep, Leshem Choshen, Yuekai Sun, and Mikhail
  Yurochkin. 2025.
\newblock \href {https://arxiv.org/abs/2412.06540} {Sloth: scaling laws for llm
  skills to predict multi-benchmark performance across families}.
\newblock \emph{Preprint}, arXiv:2412.06540.

\bibitem[{Rajpurkar et~al.(2016)Rajpurkar, Zhang, Lopyrev, and
  Liang}]{rajpurkar-etal-2016-squad}
Pranav Rajpurkar, Jian Zhang, Konstantin Lopyrev, and Percy Liang. 2016.
\newblock \href {https://doi.org/10.18653/v1/D16-1264} {{SQ}u{AD}: 100,000+
  questions for machine comprehension of text}.
\newblock In \emph{Proceedings of the 2016 Conference on Empirical Methods in
  Natural Language Processing}, pages 2383--2392, Austin, Texas. Association
  for Computational Linguistics.

\bibitem[{Rao et~al.(2024)Rao, Jain, Alon, Goues, and Hellendoorn}]{catlm}
Nikitha Rao, Kush Jain, Uri Alon, Claire~Le Goues, and Vincent~J. Hellendoorn.
  2024.
\newblock \href {https://doi.org/10.1109/ASE56229.2023.00193} {Cat-lm training
  language models on aligned code and tests}.
\newblock In \emph{Proceedings of the 38th IEEE/ACM International Conference on
  Automated Software Engineering}, ASE '23, page 409–420. IEEE Press.

\bibitem[{Rosenfeld et~al.(2019)Rosenfeld, Rosenfeld, Belinkov, and
  Shavit}]{Rosenfeld2019ACP}
Jonathan~S. Rosenfeld, Amir Rosenfeld, Yonatan Belinkov, and Nir Shavit. 2019.
\newblock \href {https://api.semanticscholar.org/CorpusID:203592013} {A
  constructive prediction of the generalization error across scales}.
\newblock \emph{ArXiv}, abs/1909.12673.

\bibitem[{Vaswani et~al.(2017)Vaswani, Shazeer, Parmar, Uszkoreit, Jones,
  Gomez, Kaiser, and Polosukhin}]{NIPS2017_3f5ee243}
Ashish Vaswani, Noam Shazeer, Niki Parmar, Jakob Uszkoreit, Llion Jones,
  Aidan~N Gomez, \L~ukasz Kaiser, and Illia Polosukhin. 2017.
\newblock \href
  {https://proceedings.neurips.cc/paper_files/paper/2017/file/3f5ee243547dee91fbd053c1c4a845aa-Paper.pdf}
  {Attention is all you need}.
\newblock In \emph{Advances in Neural Information Processing Systems},
  volume~30. Curran Associates, Inc.

\bibitem[{Xia et~al.(2024)Xia, Malladi, Gururangan, Arora, and
  Chen}]{xia2024less}
Mengzhou Xia, Sadhika Malladi, Suchin Gururangan, Sanjeev Arora, and Danqi
  Chen. 2024.
\newblock {LESS}: Selecting influential data for targeted instruction tuning.
\newblock In \emph{International Conference on Machine Learning (ICML)}.

\bibitem[{Xie et~al.(2023)Xie, Pham, Dong, Du, Liu, Lu, Liang, Le, Ma, and
  Yu}]{xie2023doremi}
Sang~Michael Xie, Hieu Pham, Xuanyi Dong, Nan Du, Hanxiao Liu, Yifeng Lu, Percy
  Liang, Quoc~V Le, Tengyu Ma, and Adams~Wei Yu. 2023.
\newblock \href {https://openreview.net/forum?id=lXuByUeHhd} {Doremi:
  Optimizing data mixtures speeds up language model pretraining}.
\newblock In \emph{Thirty-seventh Conference on Neural Information Processing
  Systems}.

\bibitem[{Ye et~al.(2025)Ye, Liu, Sun, Zhan, Zhou, and Qiu}]{ye2025data}
Jiasheng Ye, Peiju Liu, Tianxiang Sun, Jun Zhan, Yunhua Zhou, and Xipeng Qiu.
  2025.
\newblock \href {https://openreview.net/forum?id=jjCB27TMK3} {Data mixing laws:
  Optimizing data mixtures by predicting language modeling performance}.
\newblock In \emph{The Thirteenth International Conference on Learning
  Representations}.

\bibitem[{Yu et~al.(2024)Yu, Kaur, Gupta, Brown-Cohen, Goyal, and
  Arora}]{yu2024skillmix}
Dingli Yu, Simran Kaur, Arushi Gupta, Jonah Brown-Cohen, Anirudh Goyal, and
  Sanjeev Arora. 2024.
\newblock \href {https://openreview.net/forum?id=Jf5gplvglq} {{SKILL}-{MIX}: a
  flexible and expandable family of evaluations for {AI} models}.
\newblock In \emph{The Twelfth International Conference on Learning
  Representations}.

\bibitem[{Zhuo et~al.(2025)Zhuo, Chien, Chim, Hu, Yu, Widyasari, Yusuf, Zhan,
  He, Paul, Brunner, GONG, Hoang, Zebaze, Hong, Li, Kaddour, Xu, Zhang, Yadav,
  Jain, Gu, Cheng, Liu, Liu, Wang, Lo, Hui, Muennighoff, Fried, Du, de~Vries,
  and Werra}]{zhuo2025bigcodebench}
Terry~Yue Zhuo, Vu~Minh Chien, Jenny Chim, Han Hu, Wenhao Yu, Ratnadira
  Widyasari, Imam Nur~Bani Yusuf, Haolan Zhan, Junda He, Indraneil Paul, Simon
  Brunner, Chen GONG, James Hoang, Armel~Randy Zebaze, Xiaoheng Hong, Wen-Ding
  Li, Jean Kaddour, Ming Xu, Zhihan Zhang, Prateek Yadav, Naman Jain, Alex Gu,
  Zhoujun Cheng, Jiawei Liu, Qian Liu, Zijian Wang, David Lo, Binyuan Hui,
  Niklas Muennighoff, Daniel Fried, Xiaoning Du, Harm de~Vries, and Leandro~Von
  Werra. 2025.
\newblock \href {https://openreview.net/forum?id=YrycTjllL0} {Bigcodebench:
  Benchmarking code generation with diverse function calls and complex
  instructions}.
\newblock In \emph{The Thirteenth International Conference on Learning
  Representations}.

\end{thebibliography}

\onecolumn

\appendix

\section{Architecture details}
\label{app:arch_details}
All of the models used in our experiments are based on the Transformer architecture~\cite{NIPS2017_3f5ee243}, using a similar configuration to that used by \citet{llama3}. 
We scale the models to along various axes to increase the parameter counts: embedding dimension, number of attention heads, and number of layers. 
The number of layers that we use in our models ranges from 8 to 50, the number of heads ranges from 8 to 40, and the embedding dimensions range from 512 to 5120.

\section{Model pretraining}
\label{app:model_pretraining}
We follow the training recipe described in \citet{llama3} for their scaling law experiments. 
Specifically, we use a cosine learning rate schedule with 2000 steps of linear warmup from 0 and a maximum learning rate of $\eta \in [2 \times 10^{-4}, 4 \times 10^{-4}]$, which is tuned for different model sizes, that decays to a final value of $\frac{\eta}{10}$. 
Again following \citet{llama3}, we use a weight decay rate of $0.1$. 
Similarly, we vary batch sizes across compute scales from 250K to 4M tokens, keeping it constant within each scale. 

\section{Definitions}
\label{app:definitions}
Here, we provide formal definitions of the notions of IsoFLOP groups, APE COs, skill COs, and capacity- vs data-hungry skills that we discussed or introduced in \cref{sec:background}. 
\subsection{IsoFLOP groups and APE COs.}
Formally, IsoFLOP groups, which serve as the building block of IsoFLOP curves, are defined as follows. 
\begin{definition}[IsoFLOP group]
An \textit{IsoFLOP group} at a compute scale of $B$ FLOPs is an isomorphism class on the set $G_B = \{ f_t^{(p)} \in \mathcal{F} :  \text{training } f_t^{(p)} \text{for } t \text{ tokens required } B \text{ FLOPs}  \}$ with the following two isomorphisms: for a given $f_t^{(p)} \in G_B$ and decreasing $p$ to $p': 0 < p' < p$, there is a unique $f_{t'}^{(p')} \in G_B$ such that $0 < t < t'$,  and for $f_t^{(p)} \in G_B$ and increasing $p$ to $p': 0 < p < p'$, there is a unique $f_{t'}^{(p')} \in G_B$ such that $0 < t' < t$. 
\end{definition}

Next, we define the \emph{APE optimum} as the CO as measured by performance on a general validation set $\mathcal{X}_{\text{validation}}$ drawn from the same distribution as the training dataset $\mathcal{X}_{\text{train}}$, as follows. 
\begin{definition}[APE CO]
A parameter count $p^*$ and $t^*$ training tokens is called \textit{APE CO} for a dataset $\mathcal{X}_{\text{validation}}$ at a compute scale of $B$ FLOPs if $f_{t^*}^{(p^*)} \in G_B$ has improved metric $\mathcal{L}$ on samples from $\mathcal{X}_{\text{validation}}$ on average compared to any other element of $G_B$. 
\end{definition}

\subsection{Skill COs and capacity- vs data-hunger.}
Next, we move on to definitions of notions that are specifically introduced in our work. First, we define COs for skills, and then we define capacity- and data-hunger, which relate skill COs to APE COs. 

\begin{definition}[Skill CO]
    A parameter count $p^*$ and $t^*$ training tokens is called \textit{compute-optimal} for a dataset $\mathcal{D}$ that quantifies a skill $s$ at a compute scale of $B$ FLOPs if $f_{t^*}^{(p^*)} \in G_B$ has improved metric $\mathcal{L}$ on samples from $\mathcal{D}$ on average compared to any other element of $G_B$. 
\end{definition}

We call a skill capacity- or data-hungry at a particular compute scale $B$ depending on how it compares to the APE optimum as reported by \citet{llama3} for $B$. Formally, we define these as follows. 

\begin{definition}[Capacity-hungry at $B$]
A CO skill $s$ with $p_s$ parameters and $t_s$ training tokens is called \textit{capacity-hungry} at a compute scale of $B$ FLOPs if, compared to the APE CO at $B$, specified by $p_c$ and $t_c$, we have that $p_s > p_c$ and correspondingly, $t_s < t_c$. 
\end{definition}

Conversely, if a skill CO is biased toward more data than the APE CO, we have the following definition. 

\begin{definition}[Data-hungry at $B$]
A CO skill $s$ with $p_s$ parameters and $t_s$ training tokens is called \textit{data-hungry} at a compute scale of $B$ FLOPs if, compared to the APE CO at $B$, specified by $p_c$ and $t_c$, we have that $t_s > t_c$ and correspondingly, $p_s < p_c$. 
\end{definition}

\section{Evaluation dataset details}
In this section, we describe the usage of MMLU and SQuAD in our hypothesis and held-out splits.

\subsection{MMLU splits}
Our hypothesis split of MMLU~\cite{hendryckstest2021} includes Human Sexuality, Jurisprudence, Machine Learning, Marketing, Misc., Nutrition, Prehistory, Professional Psychology, Security Studies, US Foreign Policy, and World Religions, 
The held-out MMLU split contains Astronomy, Clinical Knowledge, College Chemistry, College Physics, Conceptual Physics, Electrical Engineering, High School Biology, High School European History, High School Government and Politics, High School Microeconomics, High School US History, Human Aging, International Law, Logical Fallacies, Management, Medical Genetics, Moral Disputes, Philosophy, Professional Medicine, Public Relations, Sociology, and Virology. 
We excluded all other MMLU topics, as they were either reasoning-based, or it was unclear whether they primarily evaluated \knowledge. 

\subsection{SQuAD without context} 
The SQuAD dataset~\cite{rajpurkar-etal-2016-squad} involves retrieving knowledge from a context. 
However, in this work, we are primarily interested in \knowledge-based skills that retrieve memorised facts from the training set, and not from a provided context (which may constitute an entirely different skill, such as in-context learning). 
In order to convert the SQuAD questions into a useable form for our purposes, we omit the context from our prompts and rely only on information learned during training and stored in the weights to answer the questions. 
We also discard all samples for which the answer was not contained in the context. 
We include a subsampled portion of the SquAD training set with 12000 samples in our hypothesis split and the standard SQuAD dev set in the held-out split.

\newpage

\section{Datamix ablation proportions}
\label{app:data_ablations_additional}
\begin{figure*}[h!]
\begin{subfigure}[h]{1.0\textwidth}
        \centering
	\includegraphics[width=.24\textwidth]{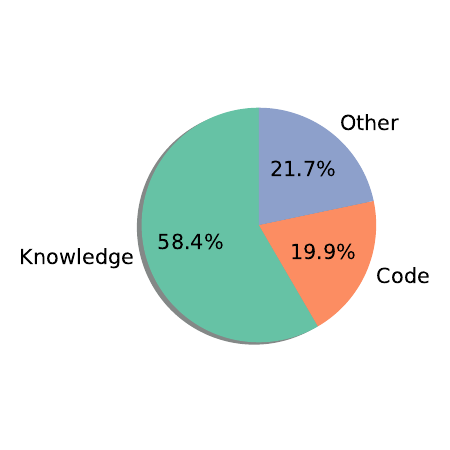}
        \caption{Canonical datamix proportions.}\label{fig:canonical_datamix_proportions}
\end{subfigure}\\
~ 
\centering
\begin{subfigure}[h]{1.0\textwidth}
        \centering
	\includegraphics[width=.19\textwidth]{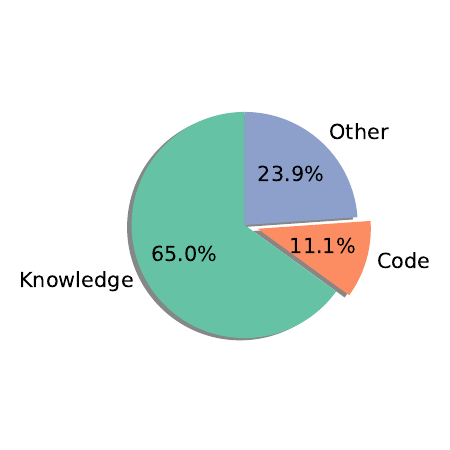}
	\includegraphics[width=.19\textwidth]{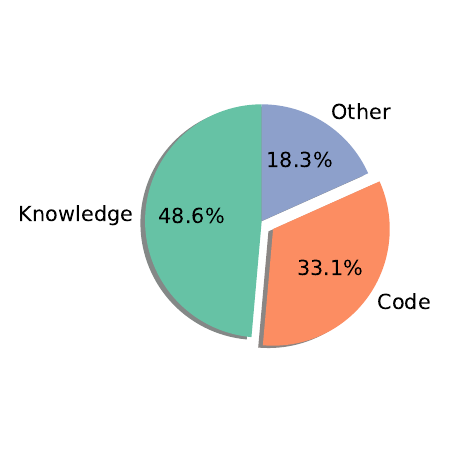}
	\includegraphics[width=.19\textwidth]{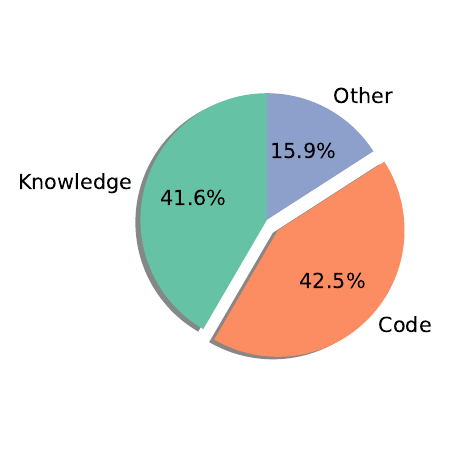}
	\includegraphics[width=.19\textwidth]{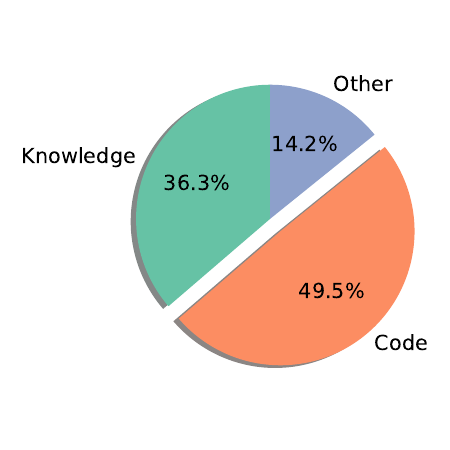}
	\includegraphics[width=.19\textwidth]{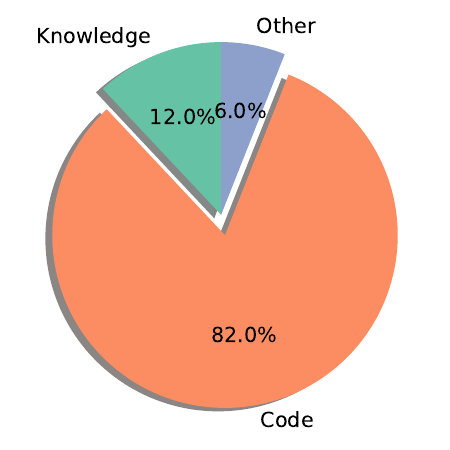}
        \caption{Code scaling proportions.}\label{fig:code_datamix_proportions}
\end{subfigure}\\
~
\begin{subfigure}[h]{1.0\textwidth}
	\centering
	\includegraphics[width=.19\textwidth]{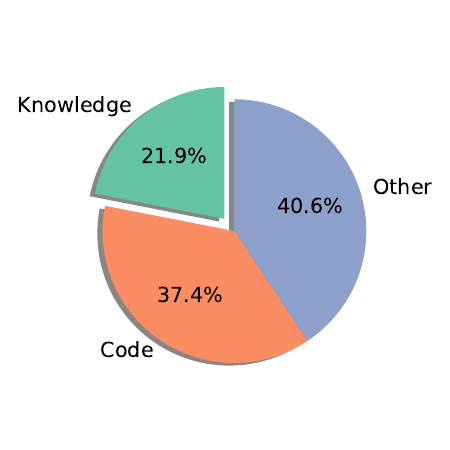}
	\includegraphics[width=.19\textwidth]{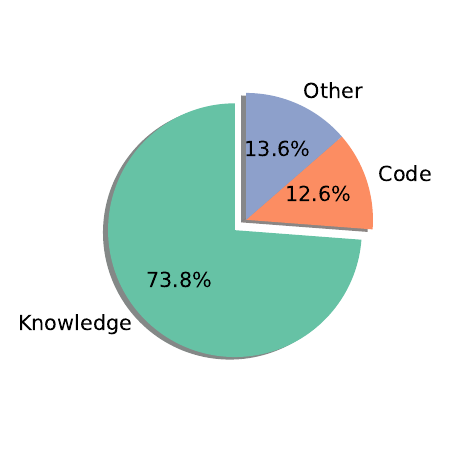}
	\includegraphics[width=.19\textwidth]{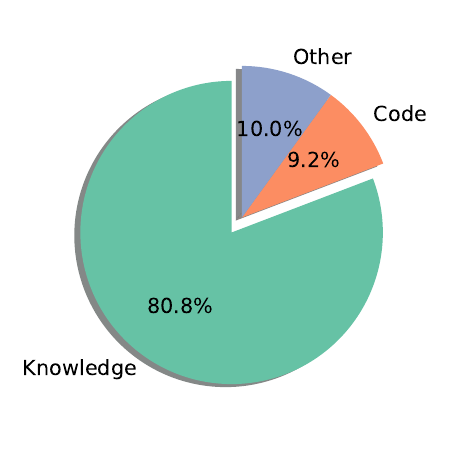}
	\includegraphics[width=.19\textwidth]{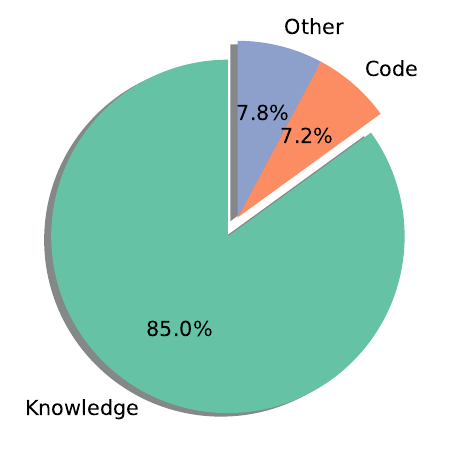}
        \caption{Knowledge scaling proportions.}\label{fig:knowledge_datamix_proportions}
\end{subfigure}
\caption{
\textbf{Pretraining datamix proportions used in our experiments.} In (a), these are the canonical datamix proportions, (b) shows the \code scaling ordered by increasing code, and in (c), our \knowledge scaling proportions ordered by increasing knowledge. 
}
\label{fig:datamix_proportions}
\end{figure*}

We provide our canonical datamix proportions in \cref{fig:canonical_datamix_proportions}, the different proportions used during \code scaling in \cref{fig:code_datamix_proportions}, and the proportions used for knowledge scaling in \cref{fig:knowledge_datamix_proportions}. 

\newpage

\section{Hypothesis split skill-relevant data scaling}
\begin{figure}[h!]
\centering
    \begin{subfigure}[t]{0.45\textwidth}
    \includegraphics[width=\columnwidth, trim=0mm 0mm 0mm 0mm, clip]{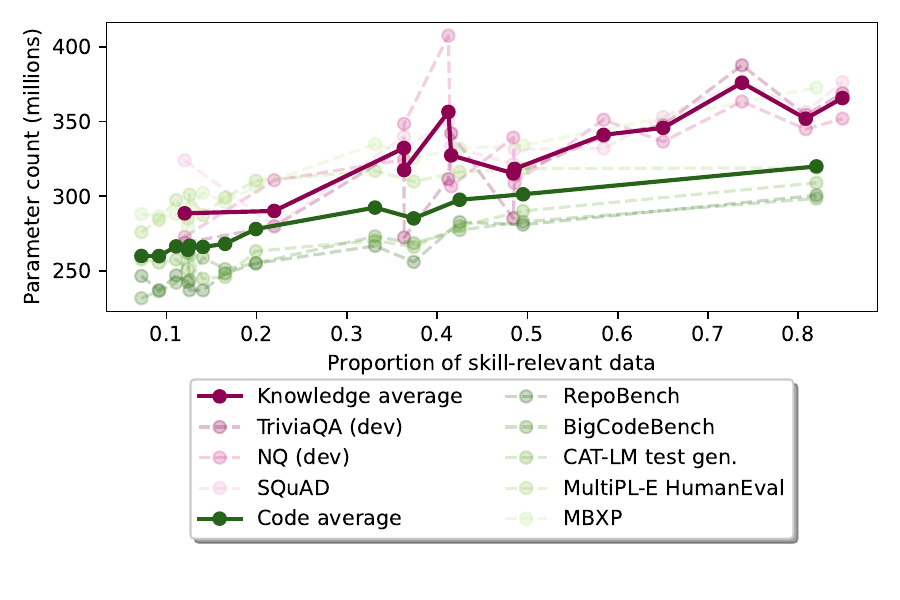}
    \caption{Hypothesis split.}
    \end{subfigure}
    \begin{subfigure}[t]{0.45\textwidth}
    \includegraphics[width=\columnwidth, trim=0mm 0mm 0mm 0mm, clip]{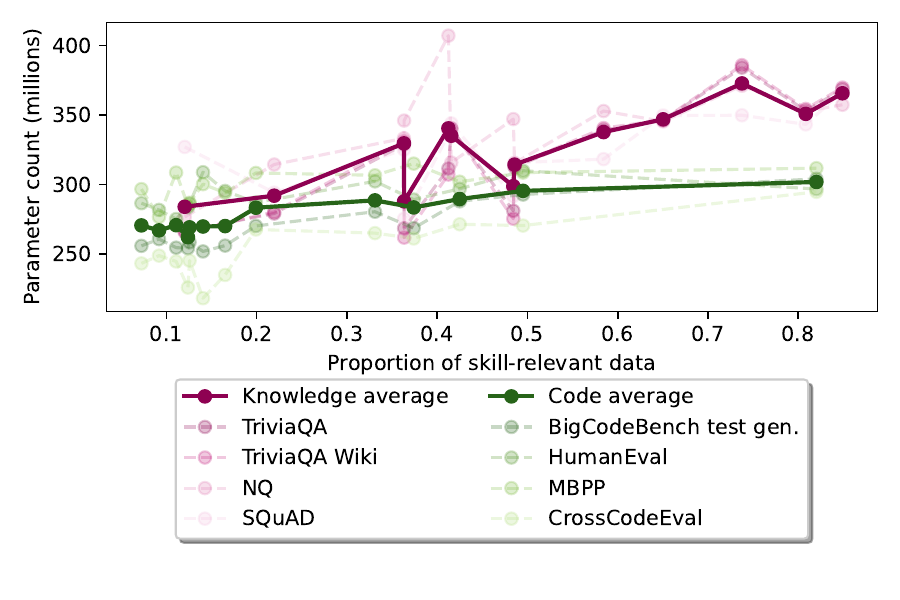}
    \caption{Held-out split.}
    \end{subfigure}
    \caption{\textbf{Comparison between hypothesis and held-out skill-relevant data scaling analyses.} For both the hypothesis and held-out splits, we find that \knowledge exhibits a faster increase in optimal parameter count as the proportion of skill-relevant data increases, implying that the capacity-hunger of \knowledge is fundamental and not an artefact of the datamix. }
    \label{fig:label_vs_proportion_hyp}
\end{figure}

In \cref{fig:label_vs_proportion_hyp}, we include a comparison of our skill-relevant data scaling analysis between the hypothesis and held-out splits. Our conclusion from the hypothesis split indeed held on the held-out split, which we reported in the main text. 

\section{Hypothesis and held-out crossover points for aligning the COs}
\label{app:crossover_comp}
\begin{figure*}[h!]
    \centering
    \begin{subfigure}[t]{0.45\textwidth}
            \includegraphics[width=\textwidth]{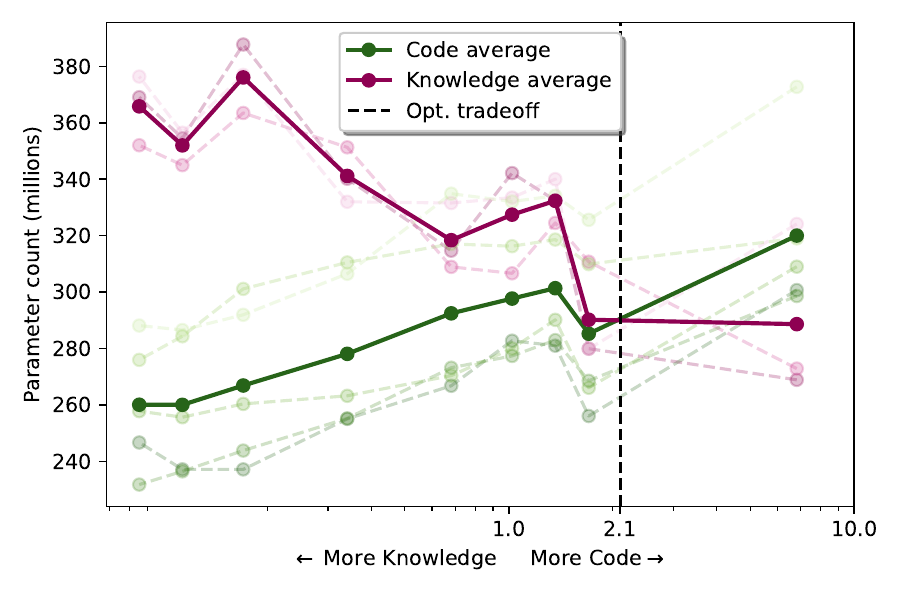}
    \end{subfigure}
    \begin{subfigure}[t]{0.45\textwidth}
            \includegraphics[width=\textwidth]{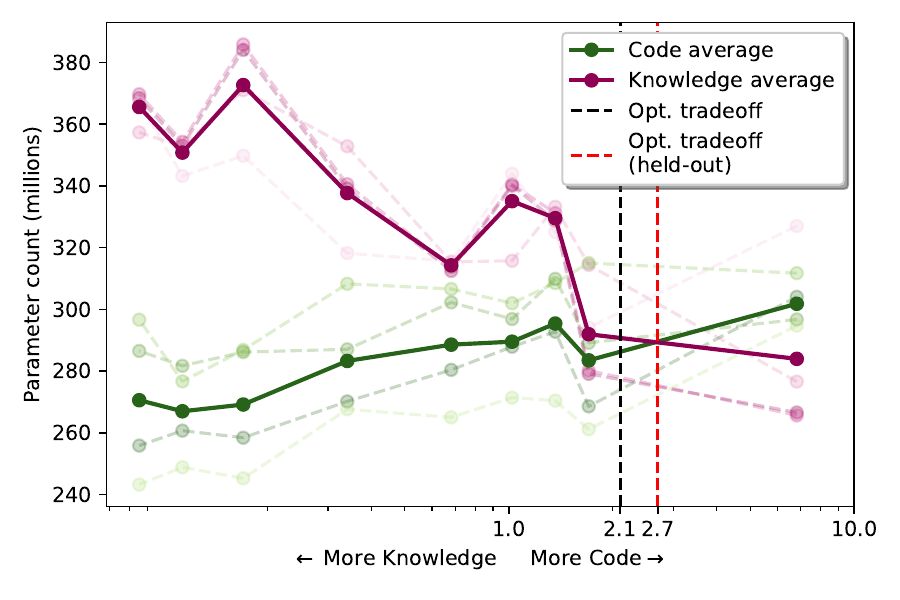}
    \end{subfigure}
    \caption{\textbf{Comparison between hypothesis set and held-out set code/knowledge scaling ratios.} We compute the crossover point that aligns the COs of the two skills on the hypothesis split (2.1) and find that the crossover point for the held-out split is similar (2.7). In the main text, we report the hypothesis set crossover point since we do not assume a priori access to the held-out set. }
\label{fig:crossover_hyp}
\end{figure*}

In addition to the held-out set results in the main text, we also include hypothesis set results for our analysis in which we computed the optimal alignment of \knowledge and \code COs in \cref{fig:crossover_hyp}. 
We note that the crossover points that align the two COs are similar between the hypothesis and held-out sets. 
We report the hypothesis crossover point in the main text to avoid a priori access to the held-out split. 

\newpage

\section{IsoFLOP curves for validation sets}
\label{app:val_isoflop}
\begin{figure*}[h!]
    \centering
    \begin{subfigure}[b]{0.11\textwidth}
            \includegraphics[height=3.8cm]{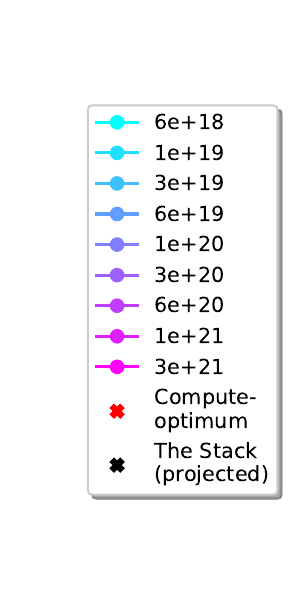}
            \vspace{1mm}
    \end{subfigure}
    \begin{subfigure}[t]{0.29\textwidth}
        \includegraphics[height=4.7cm]{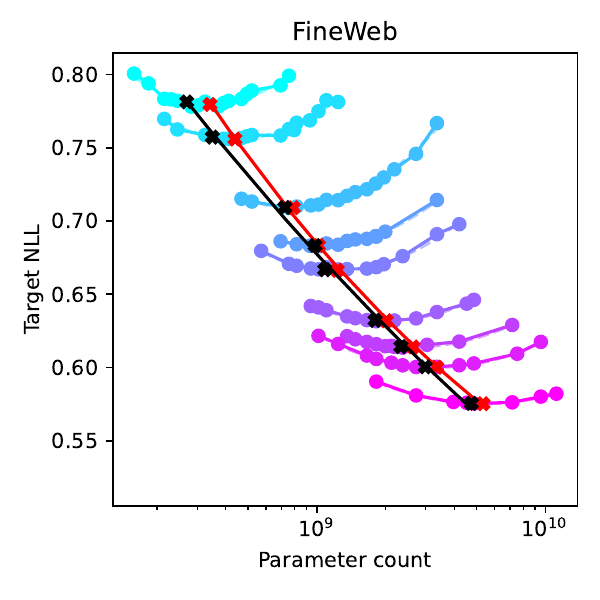}
        \caption{FineWeb validation.}
        \label{fig:fw_validation}
    \end{subfigure}
    \begin{subfigure}[t]{0.29\textwidth}
        \includegraphics[height=4.7cm]{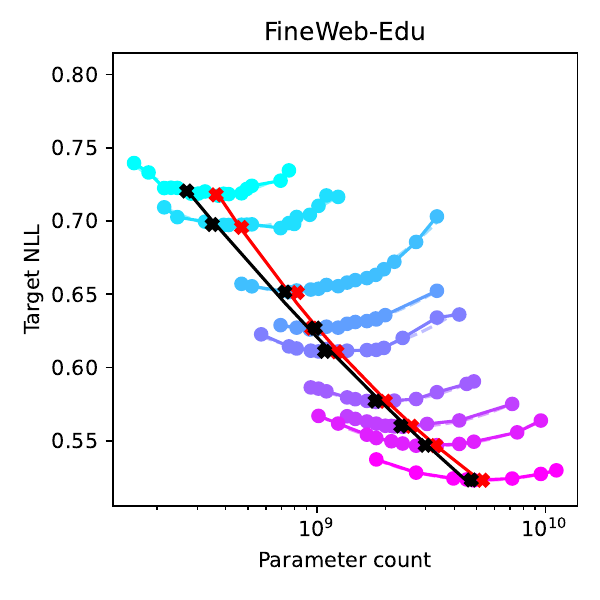}
        \caption{FineWeb-Edu validation.}
        \label{fig:fw_edu_validation}
    \end{subfigure}
    \caption{\textbf{APE COs and IsoFLOP curves of alternative validation sets.} We show the APE optima for various validation sets. All validation sets are subsampled from open source pretraining datasets, and compared to a validation set subsampled from The Stack.}
\label{fig:validation_sets}
\end{figure*}

In \cref{fig:validation_sets}, we show the IsoFLOP curves of the alternative validation sets -- FineWeb and FineWeb-Edu -- compared to The Stack. 
We find that they are all capacity-hungry compared to The Stack, which supports our findings that The Stack is more \code-aligned while the others are mode \knowledge aligned. 

\section{Per-benchmark IsoFLOP curves}
\label{app:per_benchmark_iso}
\begin{figure}[h!]
\centering
	\includegraphics[width=.45\textwidth]{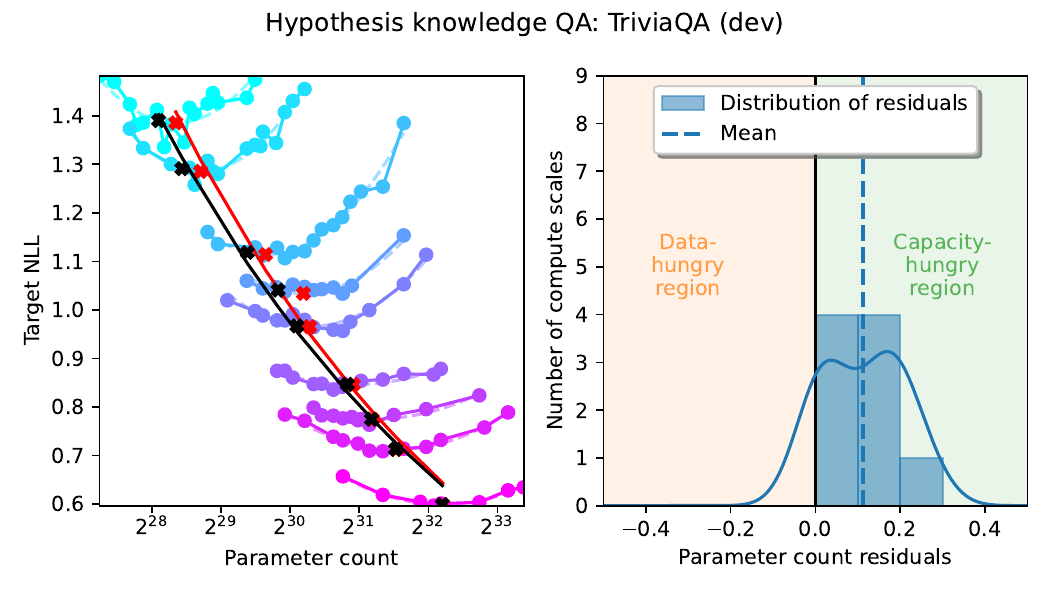}
	\includegraphics[width=.45\textwidth]{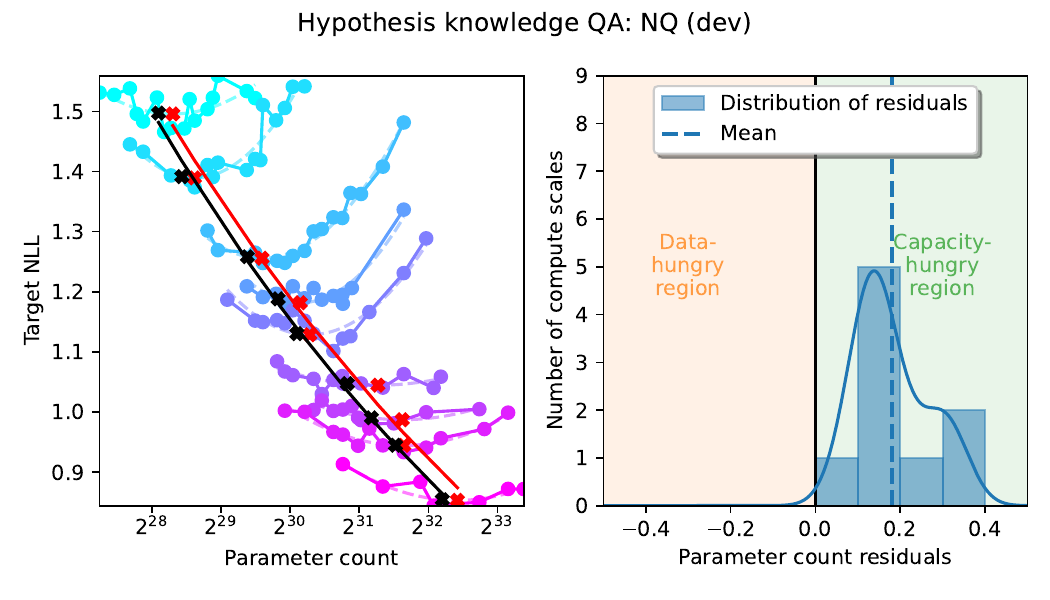} \\
	\includegraphics[width=.45\textwidth]{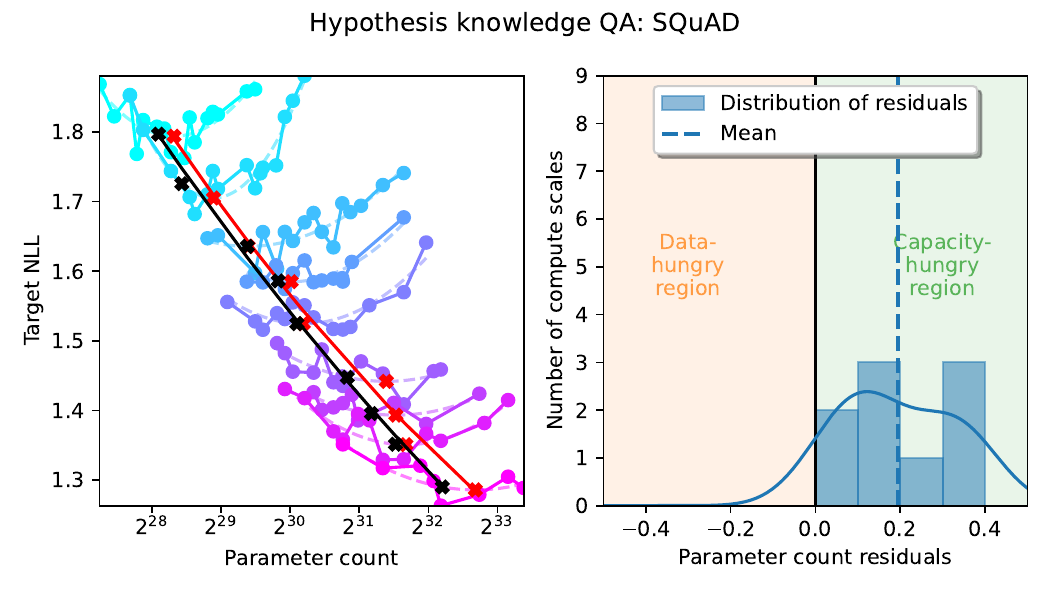}
	\includegraphics[width=.45\textwidth]{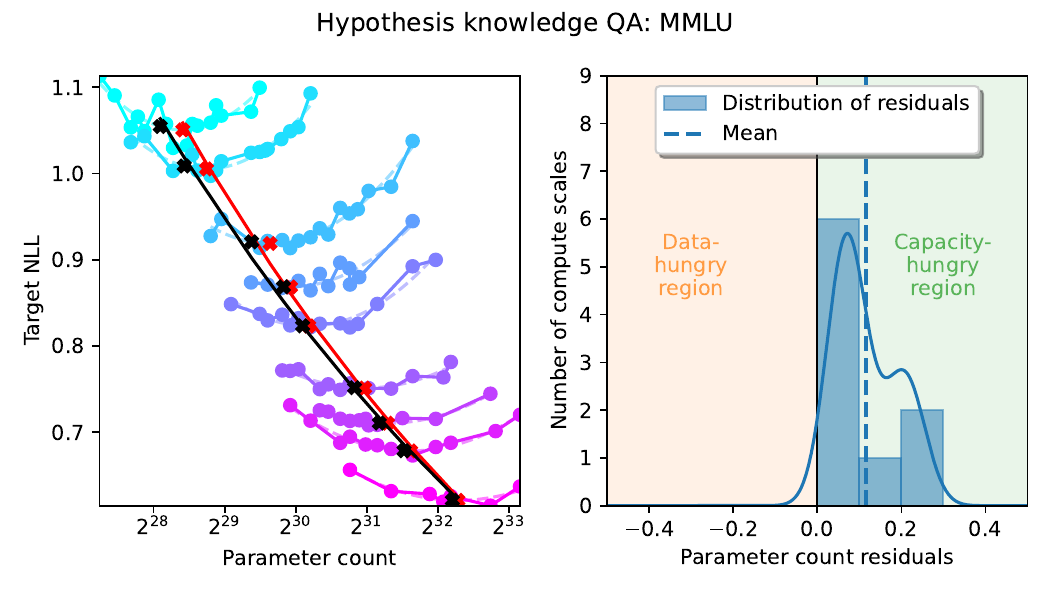}
        \caption{\textbf{Hypothesis split \knowledge skills.} On the hypothesis split, on which we formed our initial hypotheses about \knowledge vs \code scaling, we found that every \knowledge dataset exhibited capacity-hungry scaling. }\label{fig:canonical_iso_knowledge_all_dev}
\end{figure}

\begin{figure}[h!]
\centering
	\includegraphics[width=.45\textwidth]{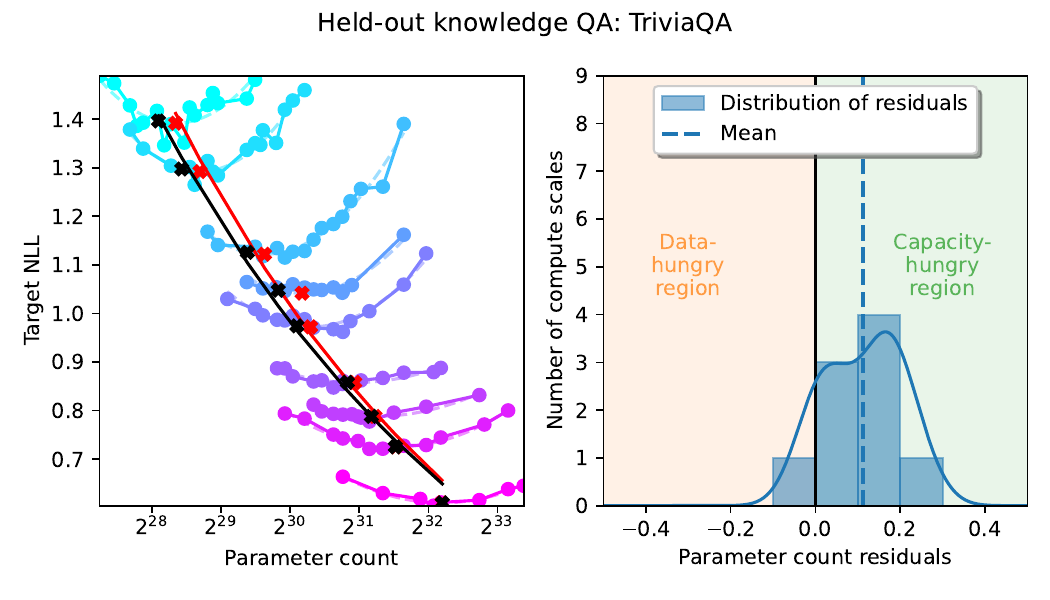}
	\includegraphics[width=.45\textwidth]{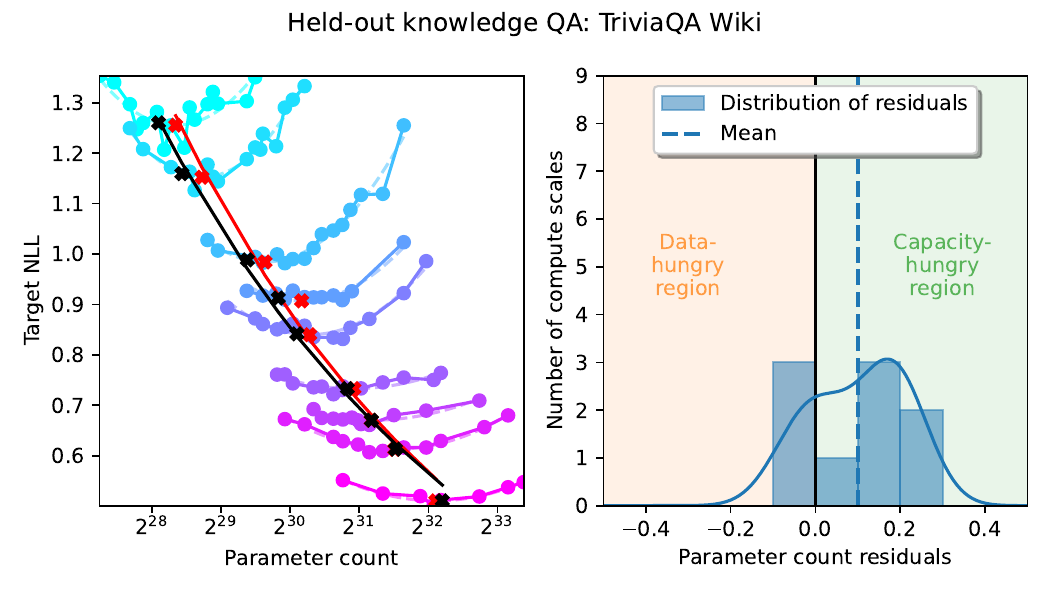} \\
	\includegraphics[width=.45\textwidth]{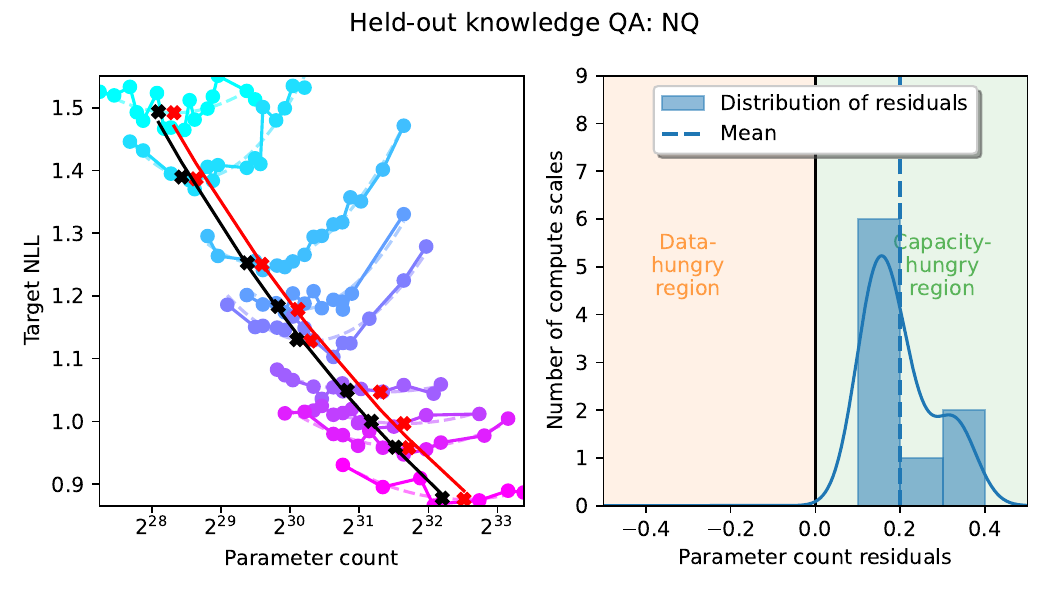}
	\includegraphics[width=.45\textwidth]{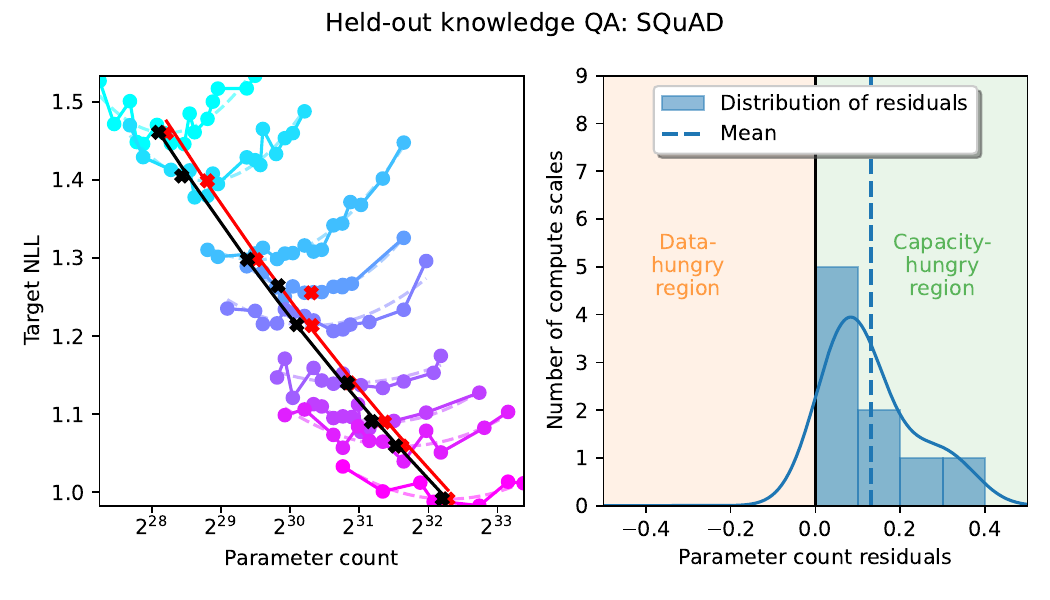} \\
         \includegraphics[width=.13\textwidth]{figs/datamix_canonical/averages_heldout/scaling_legend.pdf}
	\includegraphics[width=.45\textwidth]{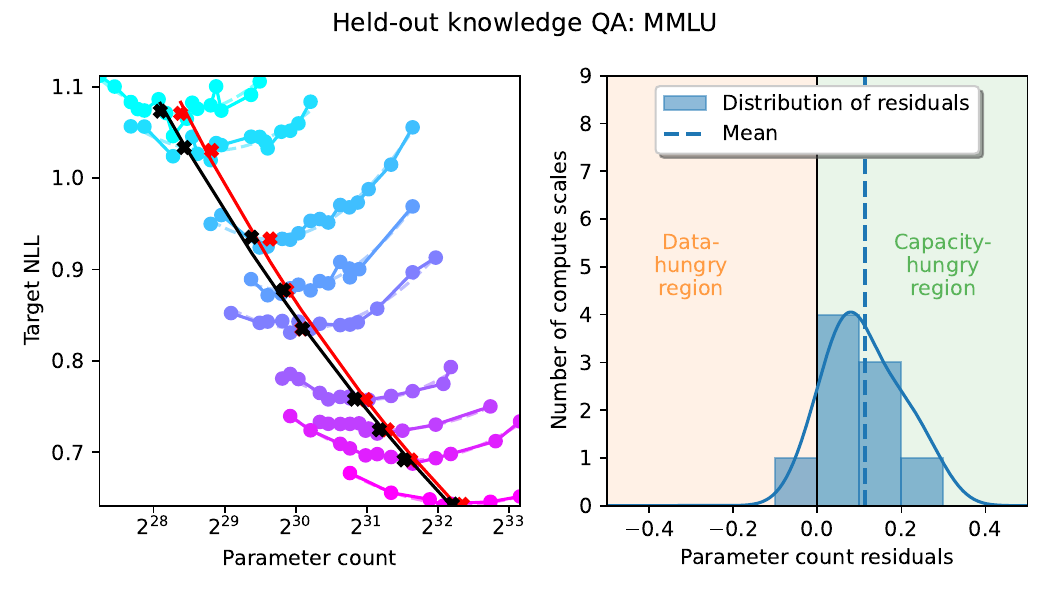} 
        \caption{\textbf{Held-out split \knowledge skills.} On the held-out split, which we did not access while forming our hypotheses, we found that every \knowledge dataset also exhibited capacity-hungry scaling.}\label{fig:canonical_iso_knowledge_all_heldout}
\end{figure}

\begin{figure}[h!]
\centering
	\includegraphics[width=.45\textwidth]{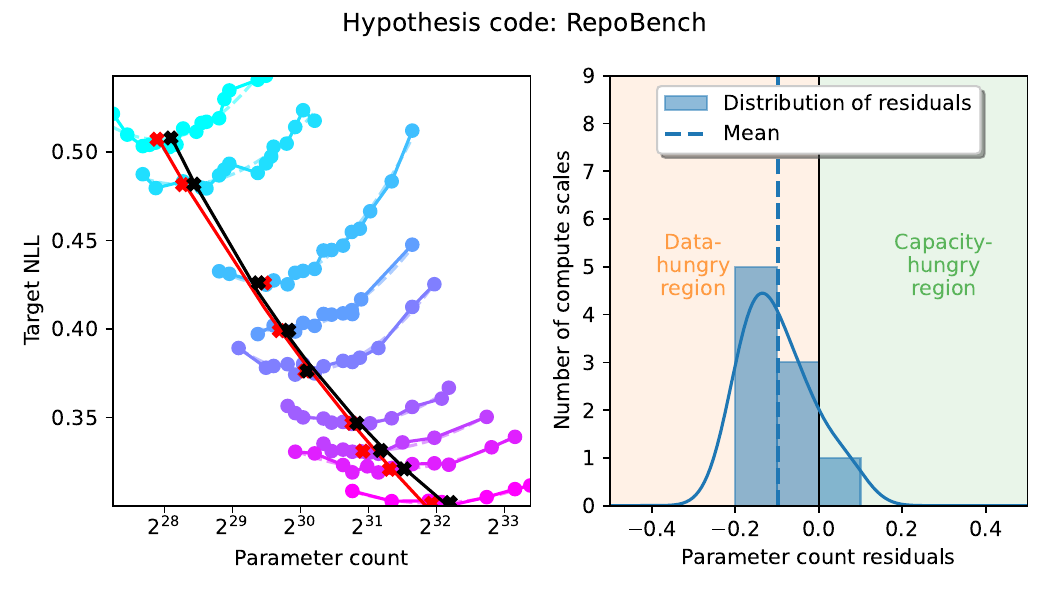}
	\includegraphics[width=.45\textwidth]{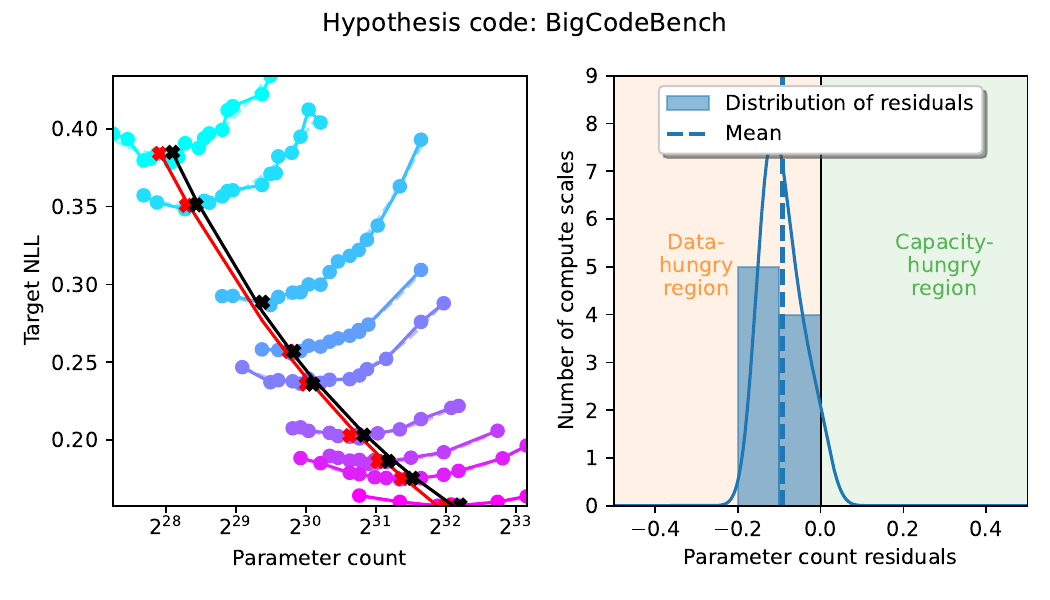} \\
	\includegraphics[width=.45\textwidth]{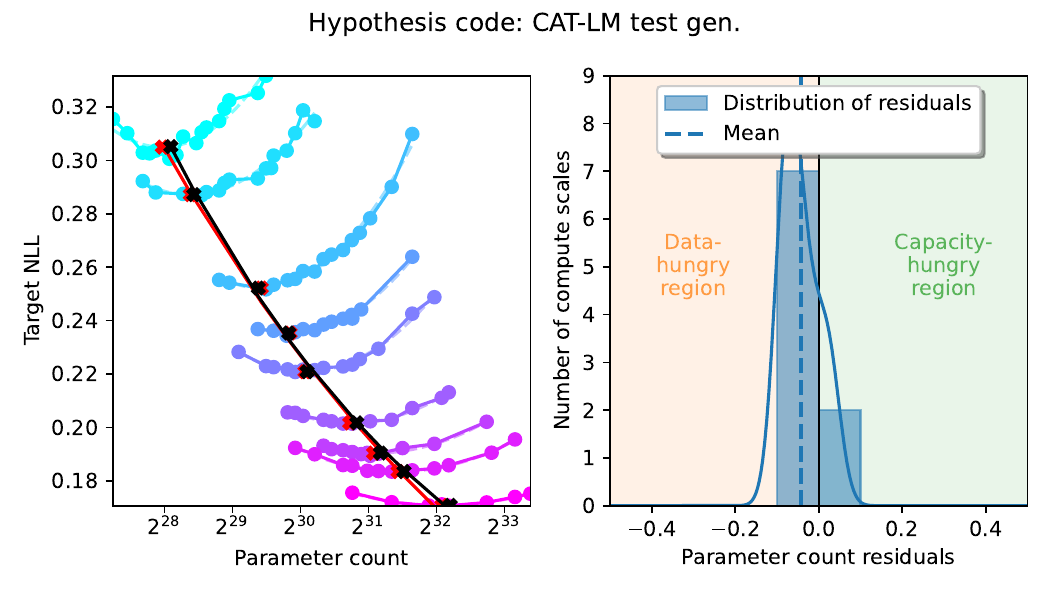}
	\includegraphics[width=.45\textwidth]{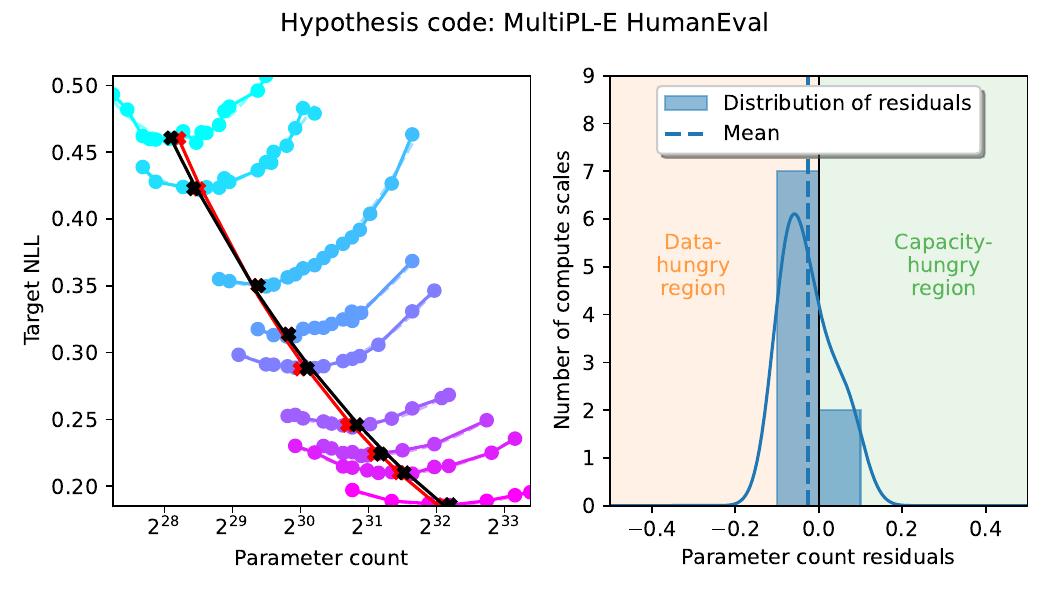} \\
         \includegraphics[width=.13\textwidth]{figs/datamix_canonical/averages_heldout/scaling_legend.pdf}
	\includegraphics[width=.45\textwidth]{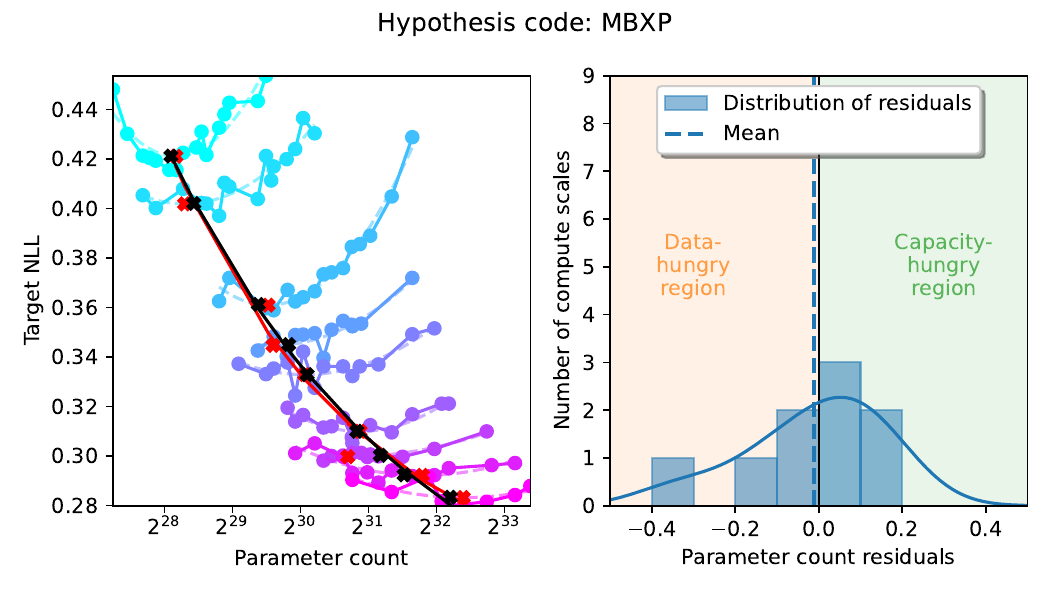} 
        \caption{\textbf{Hypothesis split \reasoning skills.} On the hypothesis split, on which we formed our initial hypotheses about \knowledge vs \code scaling, we found that every \code dataset exhibited data-hungry scaling.}\label{fig:canonical_iso_reasoning_all_dev}
\end{figure}

\begin{figure}[h!]
\centering
	\includegraphics[width=.45\textwidth]{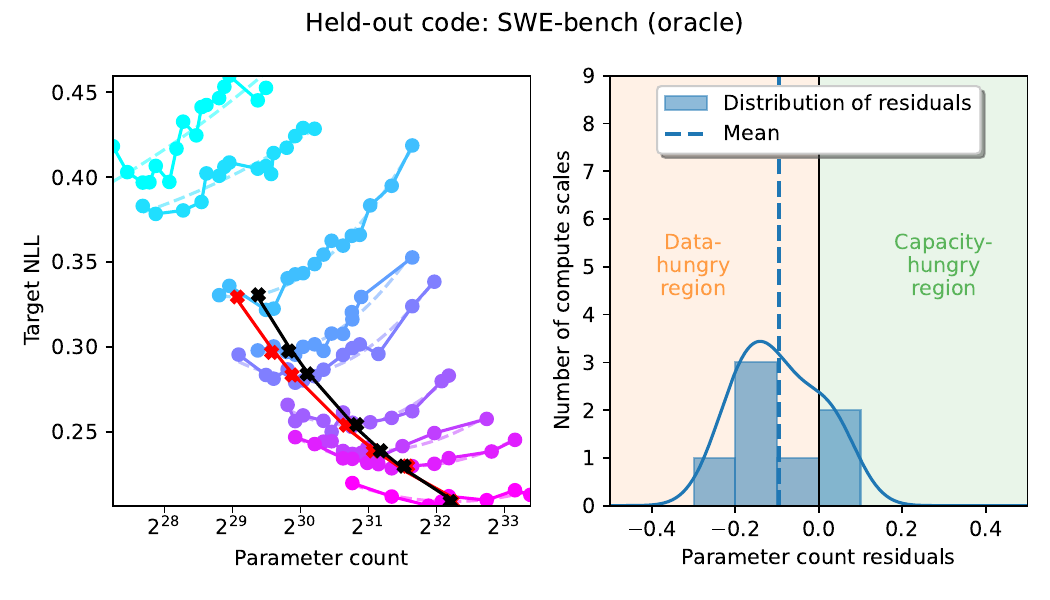}
	\includegraphics[width=.45\textwidth]{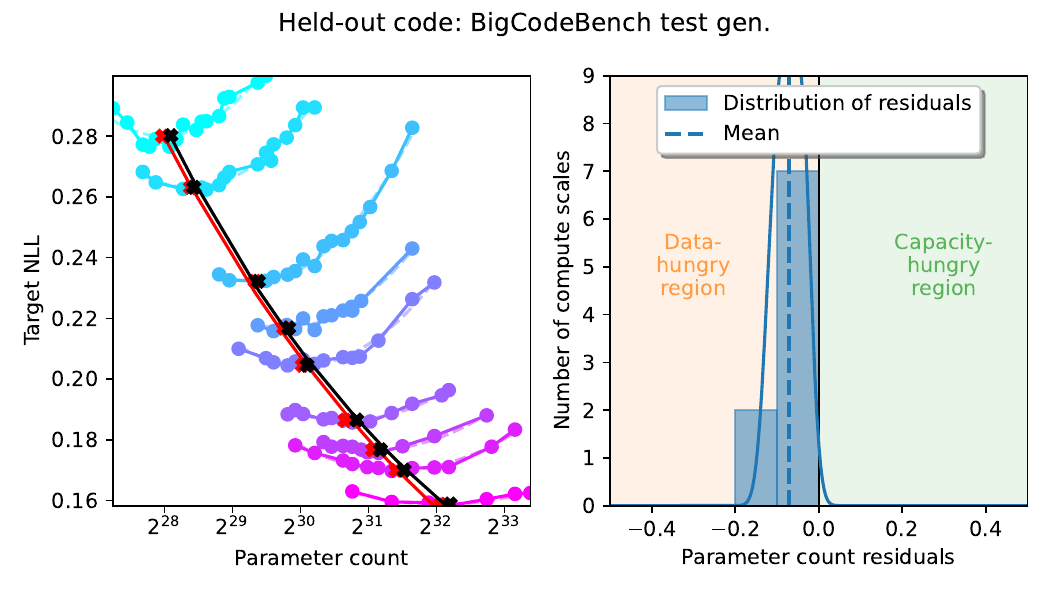} \\
	\includegraphics[width=.45\textwidth]{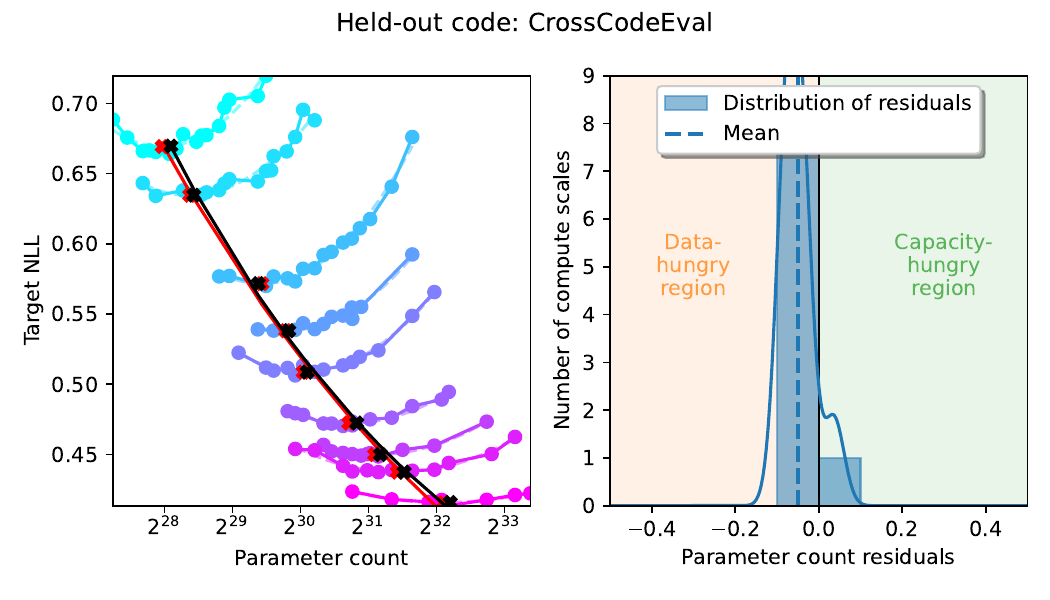}
	\includegraphics[width=.45\textwidth]{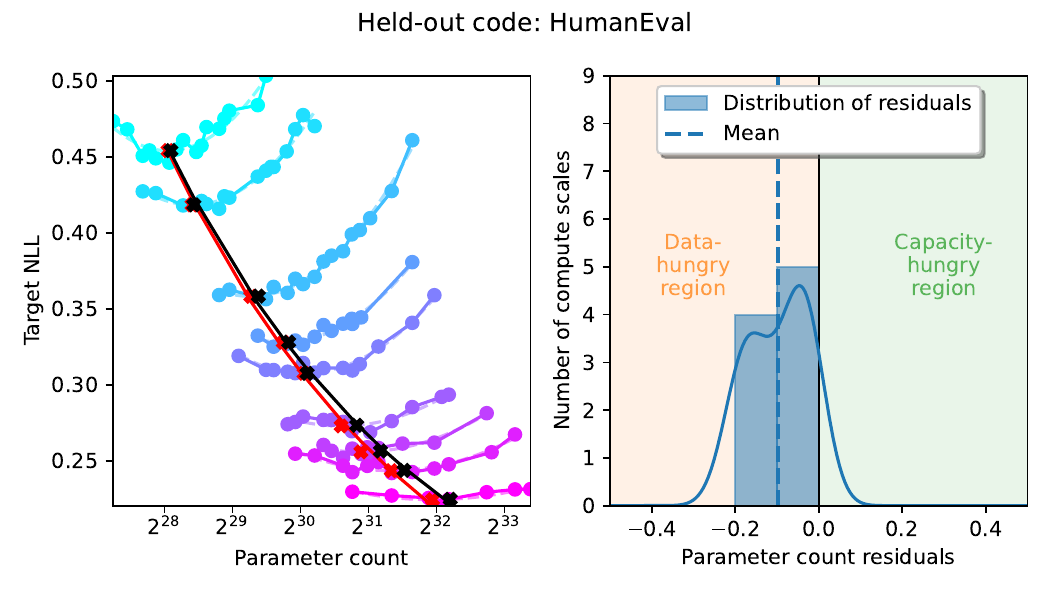} \\
         \includegraphics[width=.13\textwidth]{figs/datamix_canonical/averages_heldout/scaling_legend.pdf}
	\includegraphics[width=.45\textwidth]{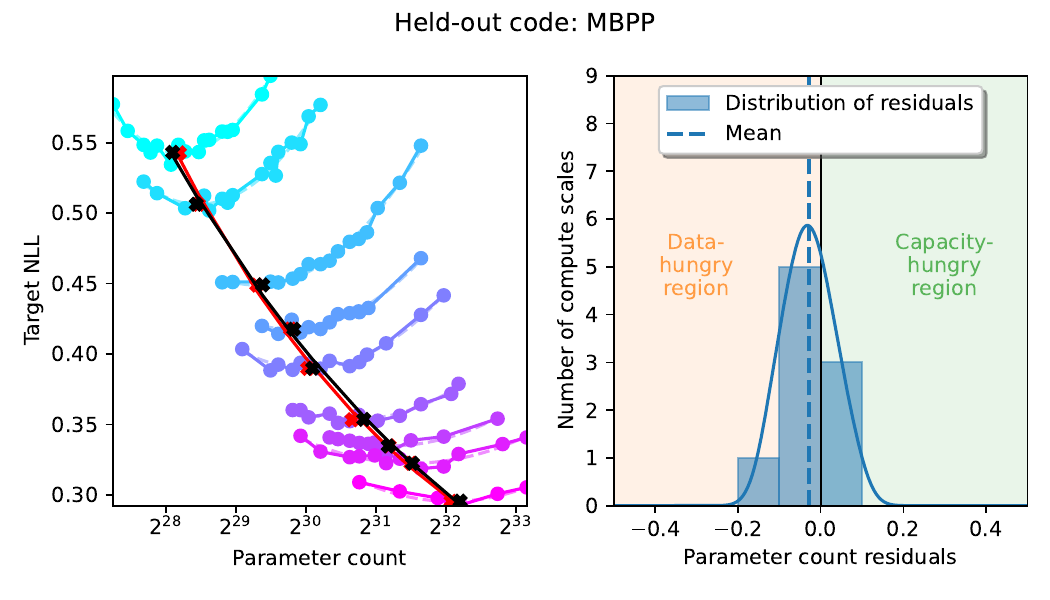} 
        \caption{\textbf{Held-out split \reasoning skills.} On the held-out split, which we did not access while forming our hypotheses, we found that every \code dataset also exhibited data-hungry scaling.}\label{fig:canonical_iso_reasoning_all_heldout}
\end{figure}

We include per-benchmark IsoFLOP curves and residual distribution plots in Figures~\cref{fig:canonical_iso_knowledge_all_dev,fig:canonical_iso_knowledge_all_heldout,fig:canonical_iso_reasoning_all_dev,fig:canonical_iso_reasoning_all_heldout}. 
Every benchmark in our hypothesis and held-out sets were capacity-hungry if they were \knowledge-based and were data-hungry if they were \code-based. 
We note that we excluded SWE-Bench (Oracle) from our final averages because its lowest two compute scales were highly skewed towards data-hunger (which is supported by SWE-Bench being \code-based), but the skew was so extreme that the optima was outside of the parameter count ranges of the models used to estimate the curves.

\section{Per-benchmark datamix ablation curves}
\label{app:per_benchmark_datamix}
\begin{figure}[h!]
\centering
	\includegraphics[width=.45\textwidth]{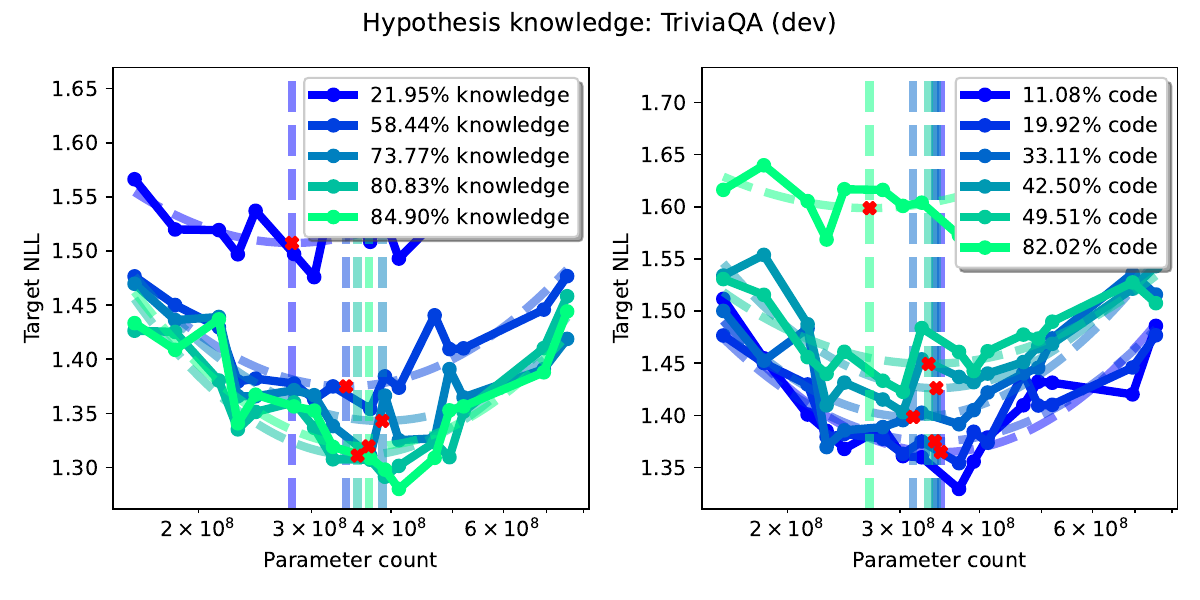}
	\includegraphics[width=.45\textwidth]{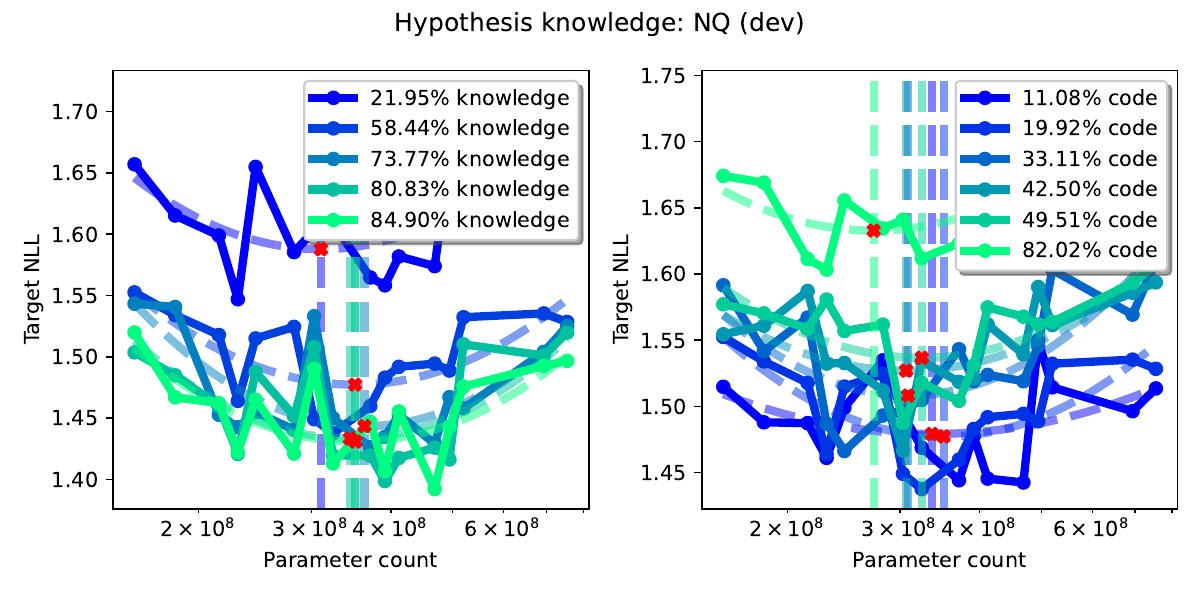} \\
	\includegraphics[width=.45\textwidth]{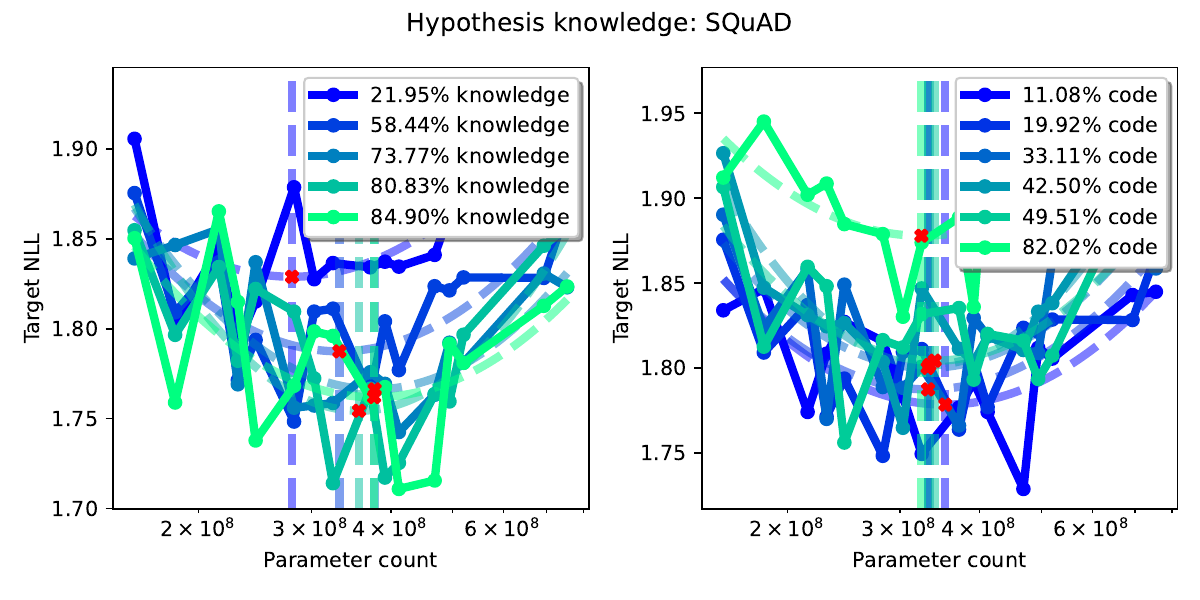}
	\includegraphics[width=.45\textwidth]{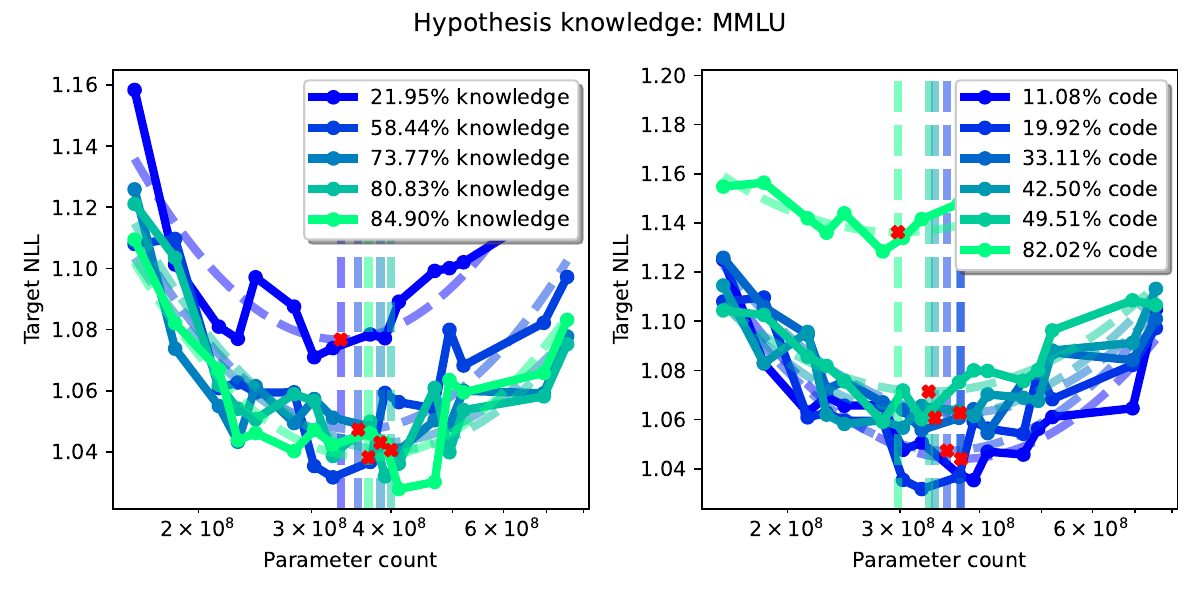}
        \caption{
        		\textbf{Data mix scaling curves for hypothesis split \knowledge skills.} For hypothesis \knowledge datasets, loss improves and COs shift to higher parameter counts with more knowledge and vice versa with code. 
        }\label{fig:datamix_ablation_knowledge_all_dev}
\end{figure}

\begin{figure}[h!]
\centering
	\includegraphics[width=.45\textwidth]{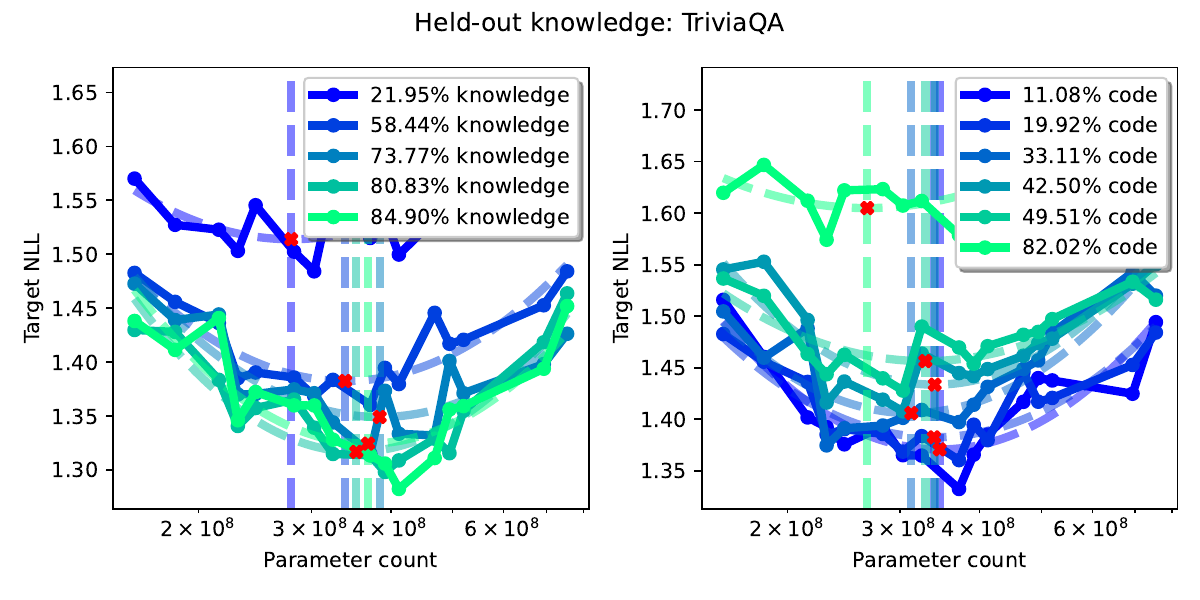}
	\includegraphics[width=.45\textwidth]{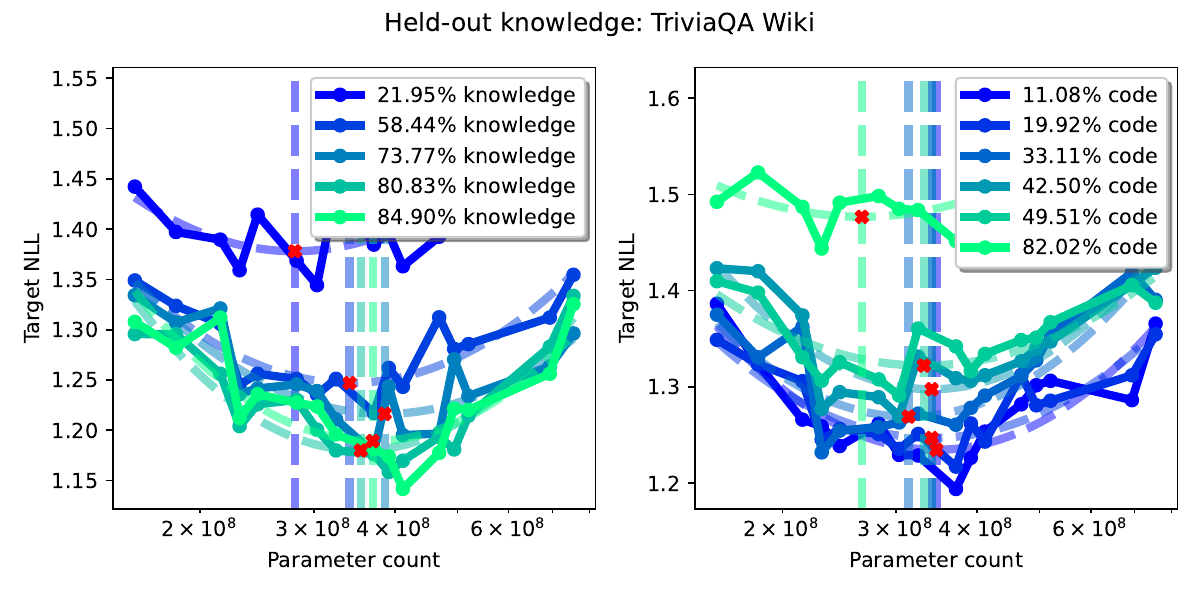} \\
	\includegraphics[width=.45\textwidth]{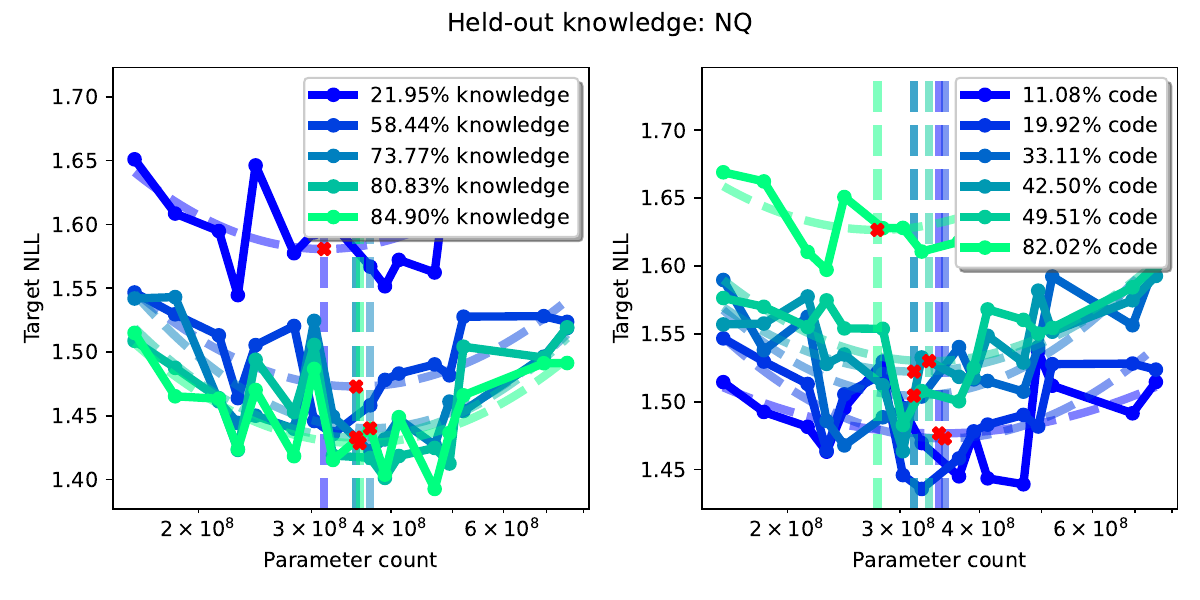}
	\includegraphics[width=.45\textwidth]{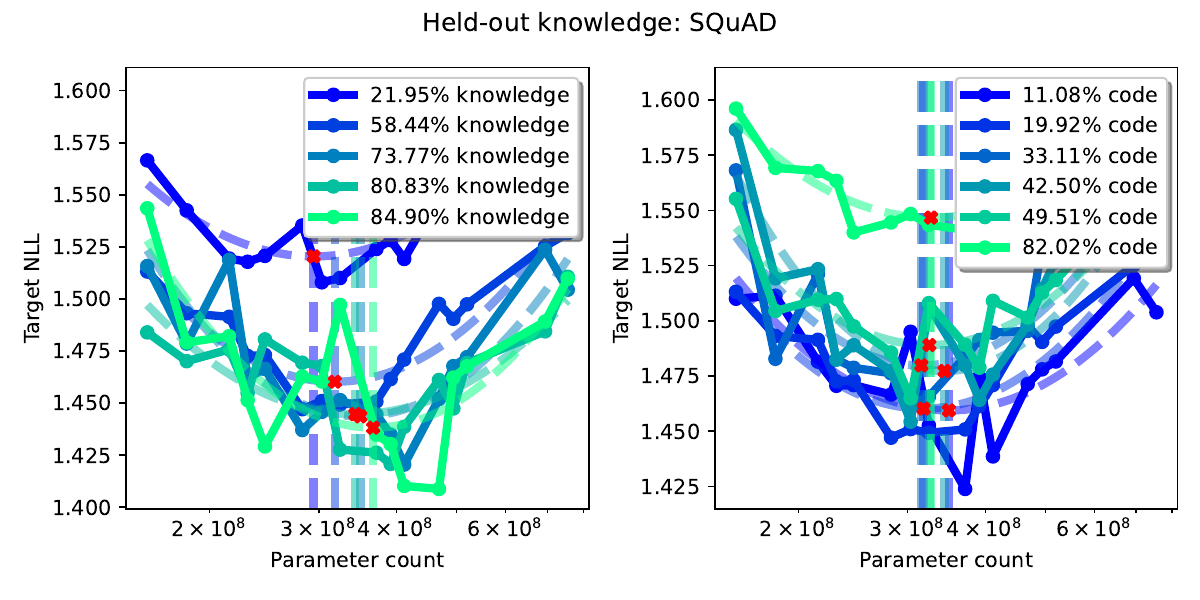} \\
	\includegraphics[width=.45\textwidth]{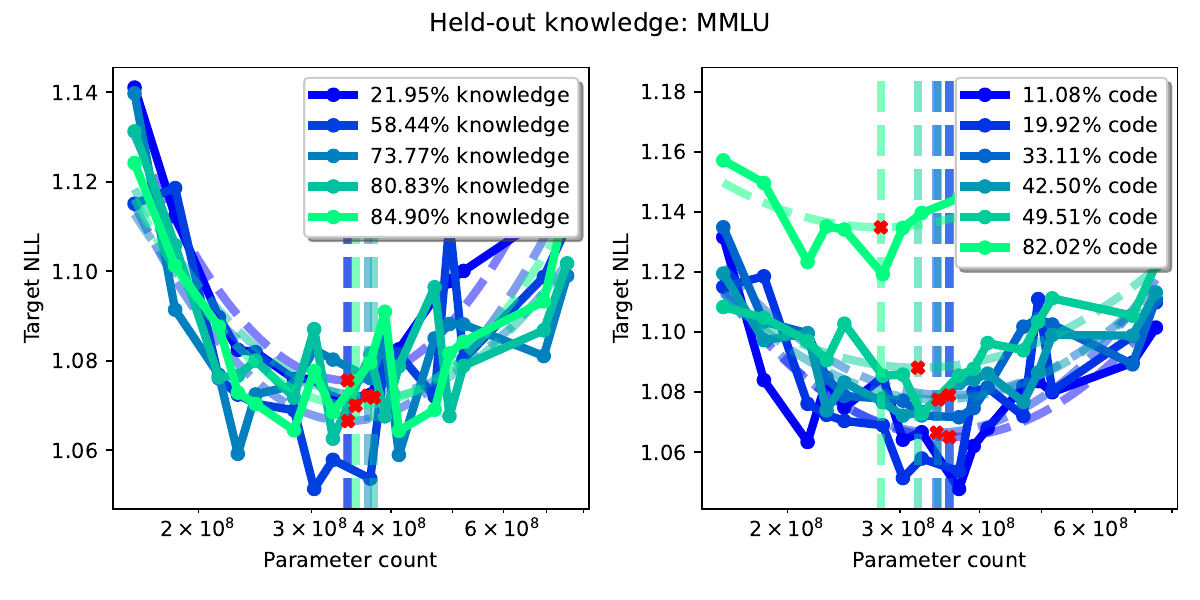} 
        \caption{
        		\textbf{Data mix scaling curves for held-out split split \knowledge skills.} For held-out \knowledge datasets except for MMLU, loss improves and COs shift to higher parameter counts with more knowledge and vice versa with more code. We attribute the noise in the MMLU results to the fact that losses are averaged across MMLU categories. 
	}\label{fig:datamix_ablation_knowledge_all_heldout}
\end{figure}

\begin{figure}[h!]
\centering
	\includegraphics[width=.45\textwidth]{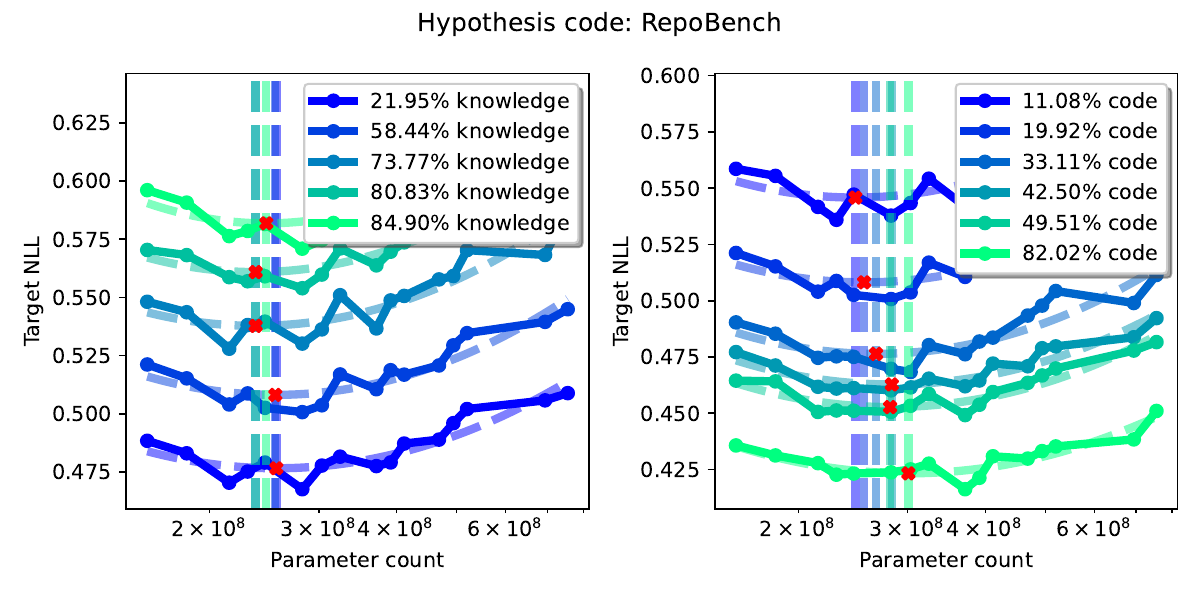}
	\includegraphics[width=.45\textwidth]{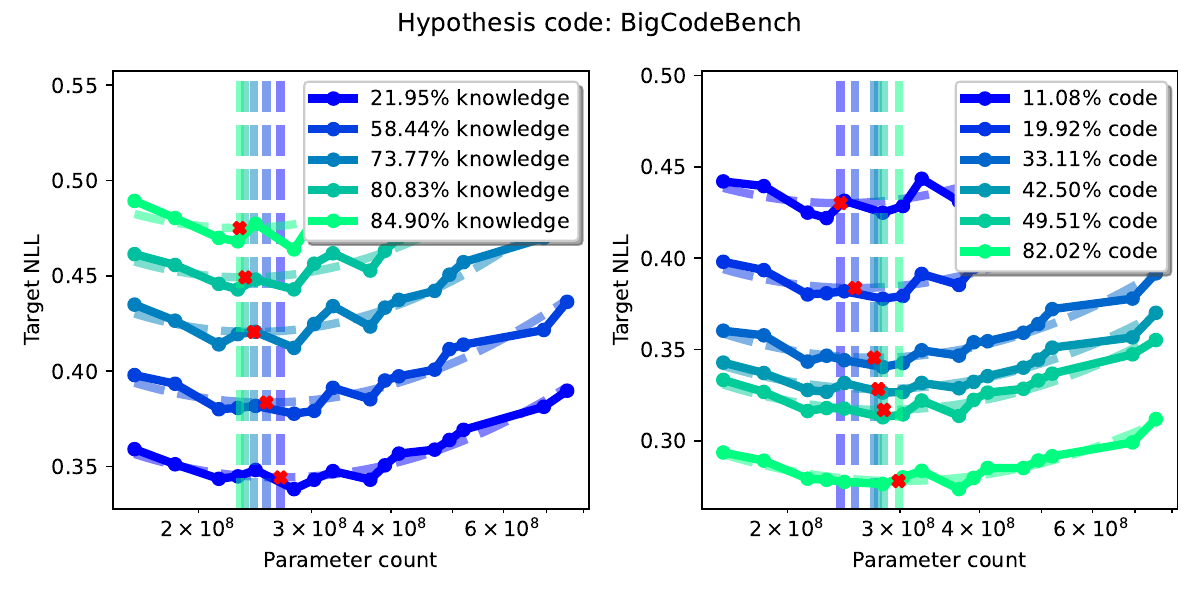} \\
	\includegraphics[width=.45\textwidth]{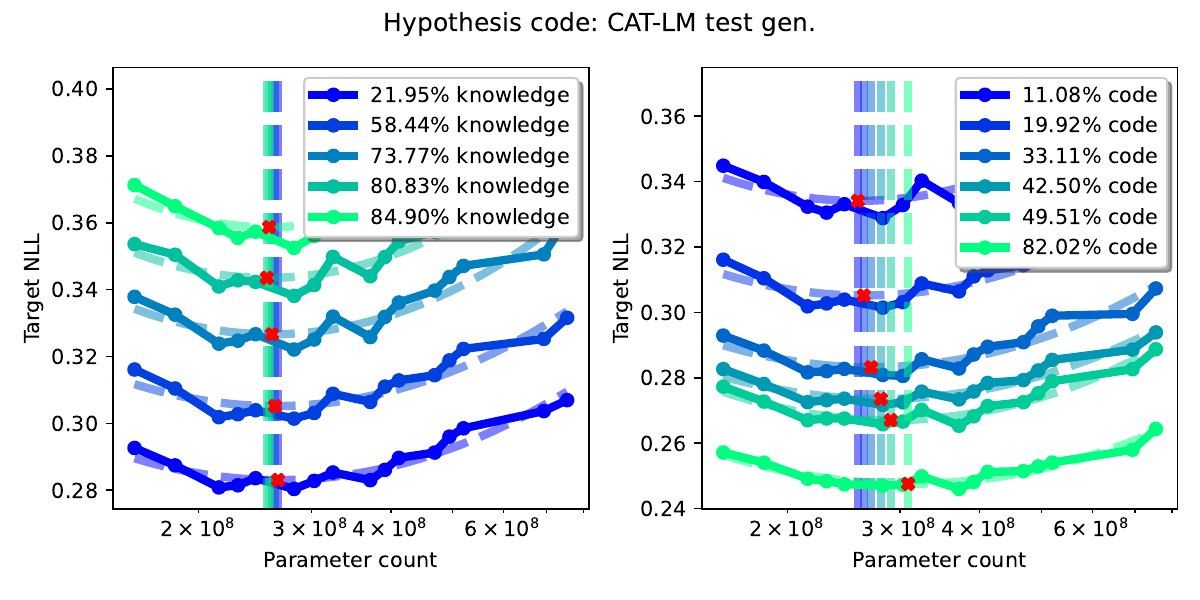}
	\includegraphics[width=.45\textwidth]{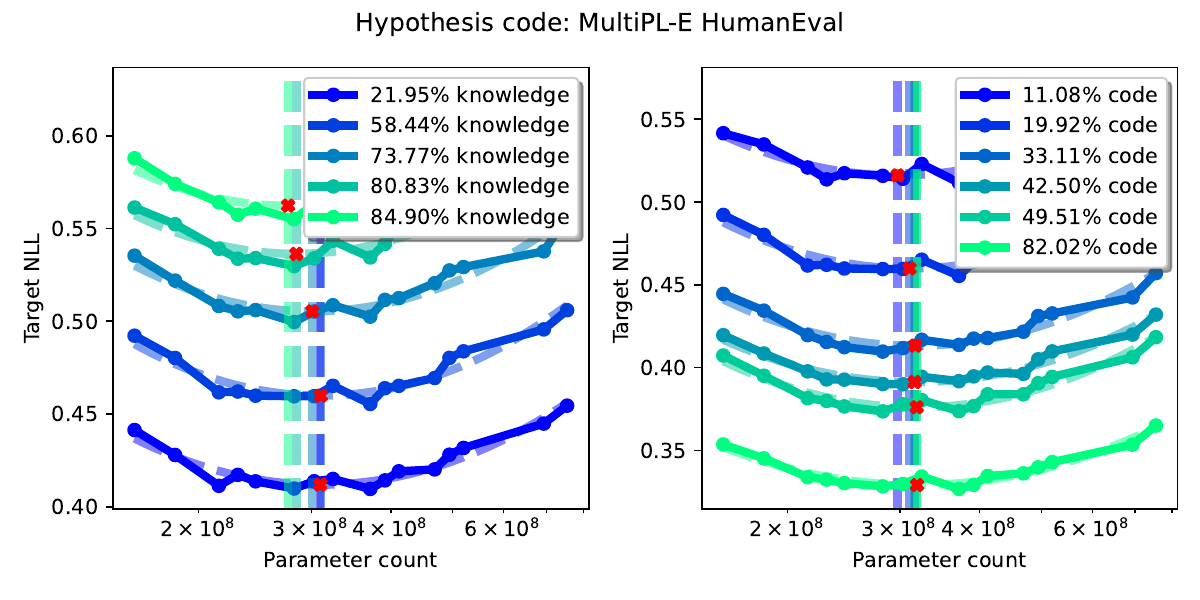} \\
	\includegraphics[width=.45\textwidth]{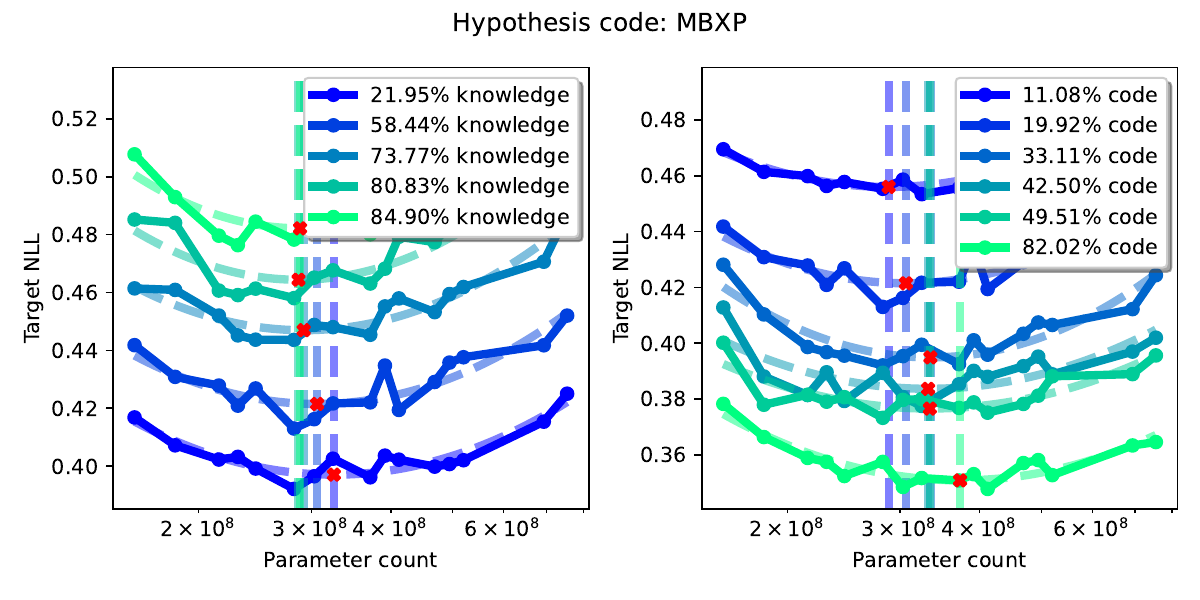} 
        \caption{
	        	\textbf{Data mix scaling curves for hypothesis split \reasoning skills.} For hypothesis \code datasets, loss improves and COs shift to higher parameter counts with more code and vice versa with knowledge. This pattern is much more clear on \code datasets than on \knowledge datasets. 
	}\label{fig:datamix_ablation_reasoning_all_dev}
\end{figure}

\begin{figure}[h!]
\centering
	\includegraphics[width=.45\textwidth]{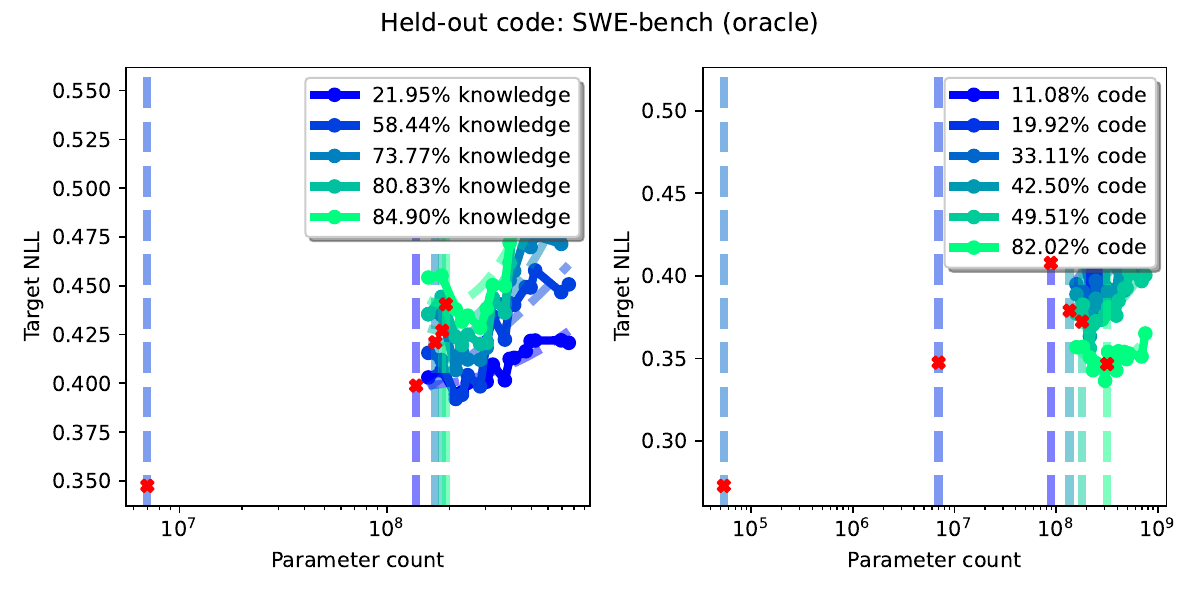}
	\includegraphics[width=.45\textwidth]{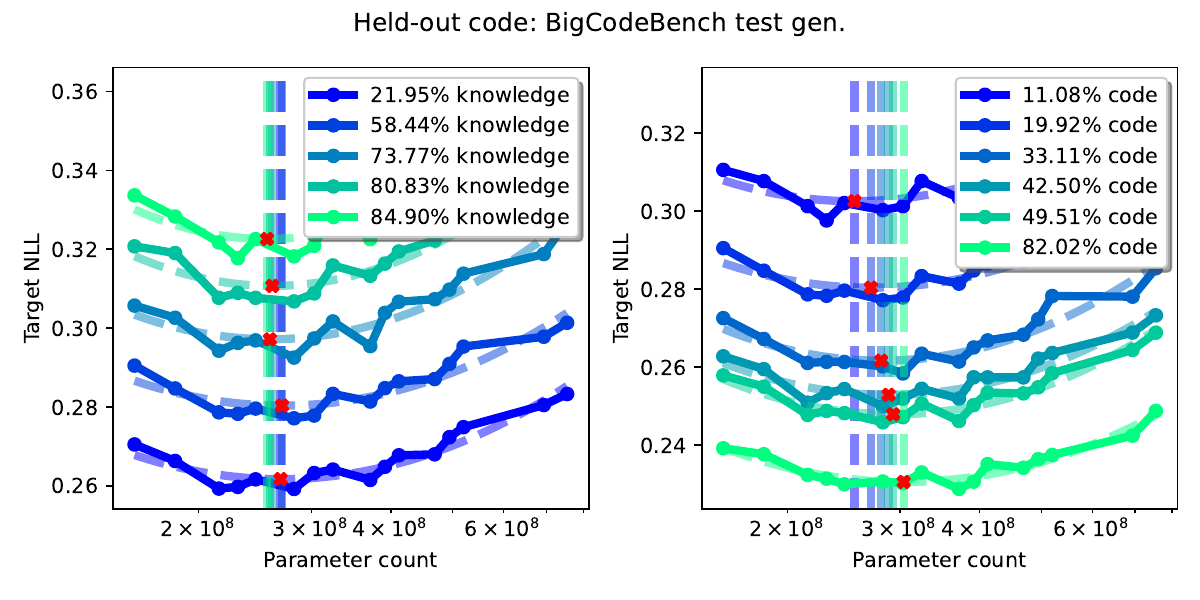} \\
	\includegraphics[width=.45\textwidth]{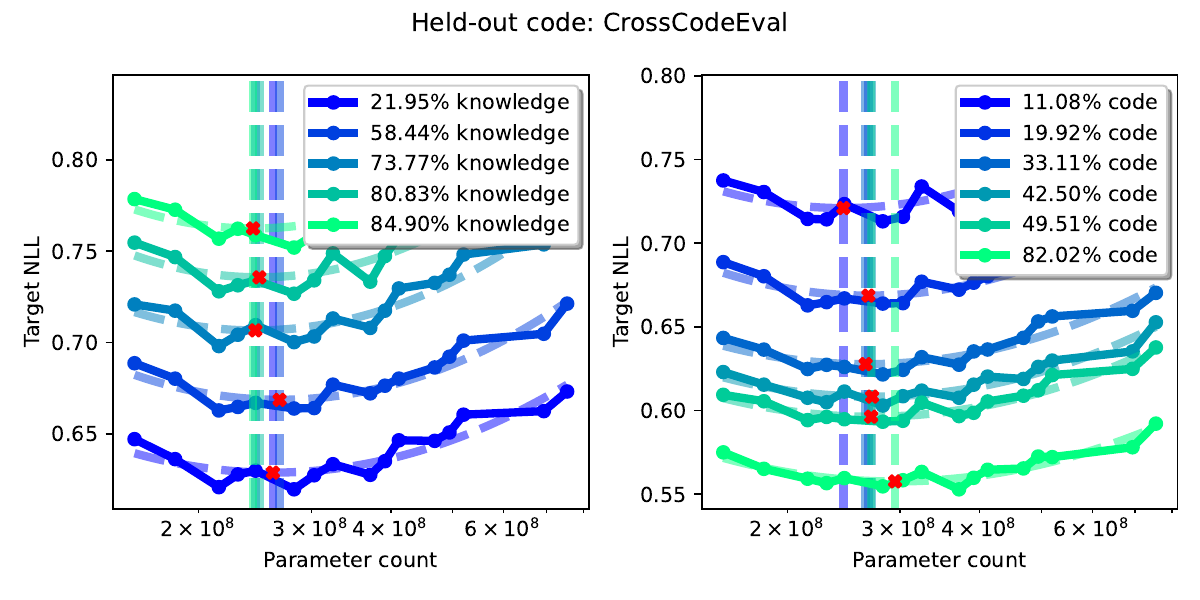}
	\includegraphics[width=.45\textwidth]{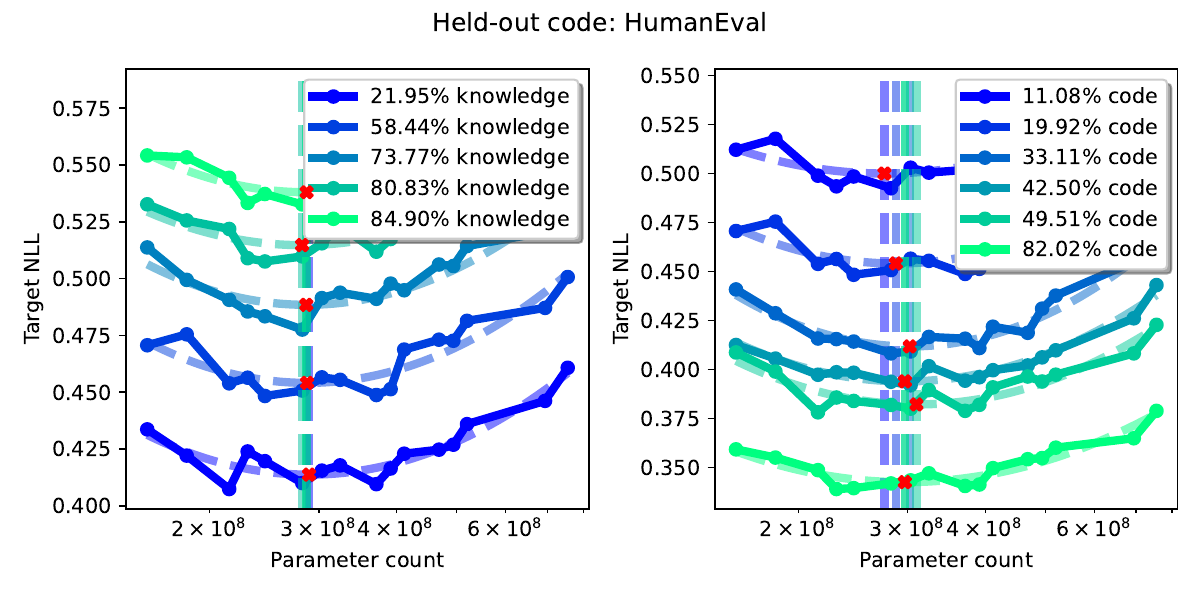} \\
	\includegraphics[width=.45\textwidth]{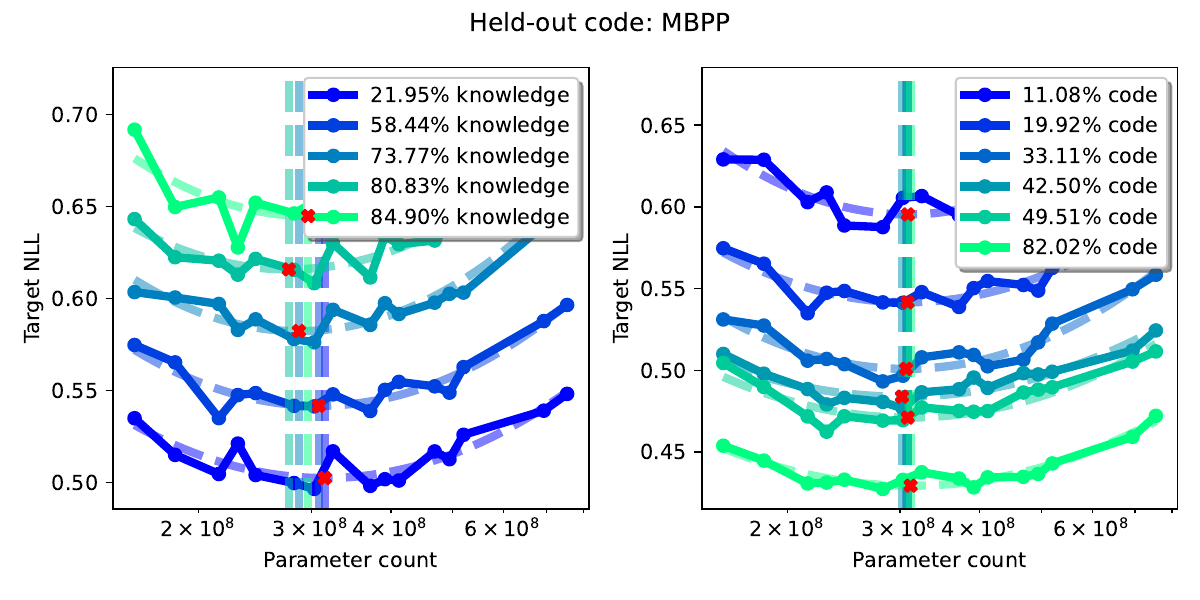} 
        \caption{
        		\textbf{Data mix scaling curves for held-out split \reasoning skills.} 
		For held-out \code datasets except for SWE-Bench (oracle), loss improves and COs shift to higher parameter counts with more code and vice versa with knowledge. 
		Note that SWE-Bench (oracle) was ultimately excluded from this analysis, as its lowest compute scale was so skewed that the estimated optima were typically outside of our empirical range. 
	}\label{fig:datamix_ablation_reasoning_all_heldout}
\end{figure}

We provide the per-benchmark plots for our datamix proportion scaling experiments in Figures~\cref{fig:datamix_ablation_knowledge_all_dev,fig:datamix_ablation_knowledge_all_heldout,fig:datamix_ablation_reasoning_all_dev,fig:datamix_ablation_reasoning_all_heldout}. 
As with our main results, we find that when we increase the proportion of \knowledge in the datamix, we see improved losses for \knowledge-based datasets and when we increase the proportion of \code in the datamix, we see improved losses for \code-based datasets, and vice versa. 
We also see that the COs shift toward capacity-hunger as we increase the proportion of skill-relevant data. 
Again, we note that we removed SWE-Bench (oracle) from the averages because the $6\times10^{18}$ compute scale was highly skewed outside of the empirical range.

\newpage

\newpage

\end{document}